\documentclass{article}

\usepackage{microtype}
\usepackage{graphicx}
\usepackage{subfigure}
\usepackage{booktabs} 

\usepackage{hyperref}

\usepackage[accepted]{icml2023}

\usepackage{amsmath}
\usepackage{amssymb}
\usepackage{mathtools}
\usepackage{amsthm}

\usepackage[capitalize,noabbrev]{cleveref}

\theoremstyle{plain}

\theoremstyle{definition}

\theoremstyle{remark}

\usepackage[textsize=tiny]{todonotes}

\usepackage{hyperref}       
\usepackage{url}            
\usepackage{booktabs}       
\usepackage{amsfonts}       
\usepackage{nicefrac}       
\usepackage{microtype}      
\usepackage{xspace}
\usepackage{paralist}

\usepackage{wrapfig}
\usepackage{amsmath}
\usepackage{amssymb}
\usepackage{amsthm}
\usepackage{url}
\usepackage{paralist}
\usepackage{latexsym}
\usepackage{arydshln}
\usepackage{tabularx}
\usepackage{verbatim}
\usepackage{CJK}
\usepackage{flushend}
\usepackage{multicol}
\usepackage{multirow,makecell}
\usepackage{color}
\usepackage{colortbl,array}
\usepackage{silence}
\usepackage{arydshln} 
\usepackage{xspace}
\usepackage{soul}
\usepackage{pifont}

\usepackage{arydshln}

\usepackage{amsmath,amsfonts,bm}

\def\eqref#1{equation~\ref{#1}}

\def\1{\bm{1}}

\def\vs{{\bm{s}}}

\DeclareMathAlphabet{\mathsfit}{\encodingdefault}{\sfdefault}{m}{sl}
\SetMathAlphabet{\mathsfit}{bold}{\encodingdefault}{\sfdefault}{bx}{n}

\newcommand{\R}{\mathbb{R}}

\DeclareMathOperator*{\argmax}{arg\,max}

\newcommand{\angstrom}{\textup{\AA}}

\usepackage{tikz}

\makeatletter
\DeclareRobustCommand\onedot{\futurelet\@let@token\@onedot}
\def\@onedot{\ifx\@let@token.\else.\null\fi\xspace}

\def\eg{\textit{e.g}\onedot} 
\def\ie{\textit{i.e}\onedot} 
 
 \def\vs{\textit{vs}\onedot}
\def\wrt{\textit{w.r.t}\onedot} 

\makeatother

\newcommand{\hidethis}[1]{}

\definecolor{bblue}{HTML}{4F81BD}
\definecolor{oorange}{HTML}{F4C842}
\definecolor{rred}{HTML}{C0504D}
\definecolor{ggreen}{HTML}{9BBB59}
\definecolor{ppurple}{HTML}{9F4C7C}
\definecolor{darkgreen}{HTML}{228B22}
\definecolor{cred}{HTML}{D81B60}
\definecolor{cblue}{HTML}{1E88E5}
\definecolor{cyellow}{HTML}{FFC107}
\definecolor{nred}{HTML}{e41a1c}
\definecolor{nblue}{HTML}{377eb8}
\definecolor{ngreen}{HTML}{4daf4a}
\definecolor{lblue}{HTML}{6C8EBF}

\renewcommand{\paragraph}[1]{\vspace{-1.1mm}\noindent\textbf{#1}}

\newlength\savewidth

\newcolumntype{x}[1]{>{\centering\arraybackslash}p{#1pt}}
\newcolumntype{y}[1]{>{\raggedright\arraybackslash}p{#1pt}}
\newcolumntype{z}[1]{>{\raggedleft\arraybackslash}p{#1pt}}

\newcommand{\app}{\raise.17ex\hbox{$\scriptstyle\sim$}}

\definecolor{deemph}{gray}{0.6}

\definecolor{baselinecolor}{gray}{.9}

\newcommand{\smallcitep}[1]{\footnotesize\citep{#1}}

\usepackage{color, colortbl}
\definecolor{emerald}{rgb}{0.31, 0.78, 0.47}
\definecolor{Gray}{gray}{0.9}
\definecolor{Highlight}{rgb}{0.89,0.89,0.94}
\usepackage[first=0,last=9]{lcg}
\newcommand{\chl}{\cellcolor{Highlight}}

\newcommand{\textbi}[1]{\textit{\textbf{#1}}}

\newcommand{\method}{\textsc{{LM-Design}}\xspace}

\newcommand{\pLM}{\textit{p}\textsc{LM}\xspace}
\newcommand{\pLMs}{\textit{p}\textsc{LM}s\xspace}

\begin{document}

\twocolumn[
\icmltitle{Structure-informed Language Models Are Protein Designers}



\icmlsetsymbol{equal}{*}

\begin{icmlauthorlist}
\icmlauthor{Zaixiang Zheng}{equal,byted}
\icmlauthor{Yifan Deng}{equal,uwm,byted}
\icmlauthor{Dongyu Xue}{byted}
\icmlauthor{Yi Zhou}{byted}
\icmlauthor{Fei Ye}{byted}
\icmlauthor{Quanquan Gu}{byted}
\end{icmlauthorlist}

\icmlaffiliation{byted}{ByteDance AI Lab}
\icmlaffiliation{uwm}{Dept. of Computer Science, University of Wisconsin-Madison (work was done during Yifan's internship at ByteDance)}

\icmlcorrespondingauthor{Zaixiang Zheng}{zhengzaixiang@bytedance.com}
\icmlcorrespondingauthor{Fei Ye}{yefei.joyce@bytedance.com}

\icmlkeywords{Machine Learning, ICML}

\vskip 0.3in
]



\printAffiliationsAndNotice{\icmlEqualContribution} 


\begin{abstract}

This paper demonstrates that language models are strong structure-based protein designers.
We present \method, a generic approach to reprogramming sequence-based protein language models (\pLMs), that have learned massive sequential evolutionary knowledge from the universe of natural protein sequences, to acquire an immediate capability to design preferable protein sequences for given folds.
We conduct a \textit{structural surgery} on \pLMs, where a lightweight structural adapter is implanted into \pLMs and endows it with structural awareness.
During inference, iterative refinement is performed to effectively optimize the generated protein sequences.
Experiments show that \method improves the state-of-the-art results by a large margin, leading to 4\% to 12\% accuracy gains in sequence recovery (\eg, 55.65\%/56.63\% on CATH 4.2/4.3 single-chain benchmarks, and $>$60\% when designing protein complexes).
We provide extensive and in-depth analyses, which verify that \method can (1) indeed leverage both structural and sequential knowledge to accurately handle structurally non-deterministic regions, (2) benefit from scaling data and model size, and (3) generalize to other proteins (\eg, antibodies and \textit{de novo} proteins).

\end{abstract}

\vspace{-8mm}
\section{Introduction}
\label{sec:intro}

{Proteins are 3D-folded linear chains of amino acids that govern biological functions such as transcription, translation, signaling, and cell cycle control.}
Recently, the promise of learning to understand and design proteins from data via generative deep learning has led to an ongoing paradigm shift apart from the long-established physics-based methods.

{Designing protein sequences that fold into desired structures, namely structure-based protein (sequence) design, is one of the most important problems in bio-engineering.}
Significant progress has been made by several latest deep generative model-based approaches ~\citep{ingraham2019generative,jing2020gvp,hsu2022esmif,dauparas2022proteinmpnn,gao2022pifold}. 
These approaches formulate structure-based protein design as an end-to-end graph-to-sequence learning problem, where an encoder-decoder model $\mathcal{M}_{\theta}: \mathcal{X} \to \mathcal{S}$ is tasked with predicting protein sequence $\mathcal{S}$ given a protein backbone structure $\mathcal{X}$.
Typically, supervised learning is performed on such models given a certain amount of protein structure-sequence pair data.

\begin{figure*}[t]
    \vspace{-1.5mm}
    \centering
    \includegraphics[width=0.995\linewidth]{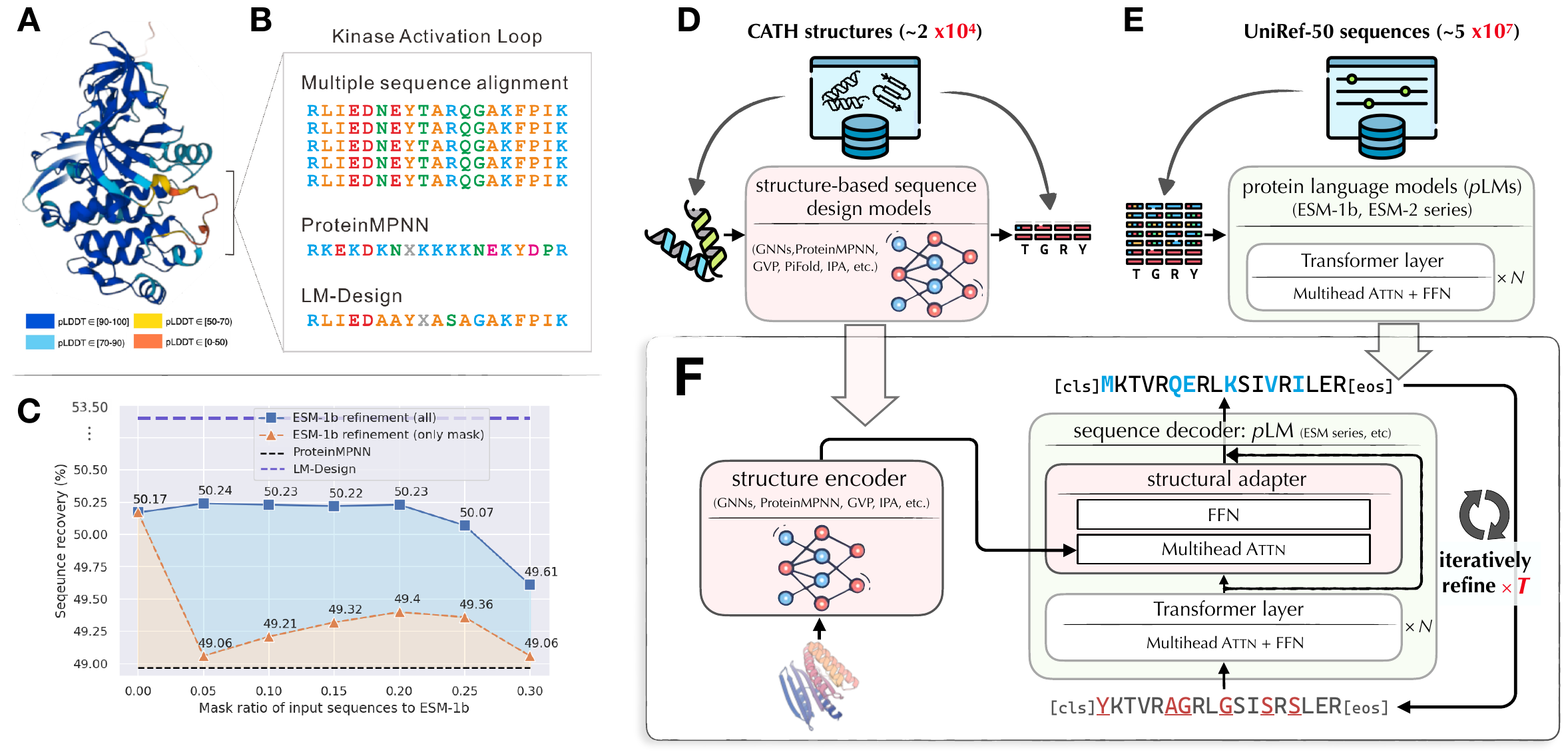}
    \vspace{-4.5mm}
    \caption{\emph{Overview.}
    \textbf{(A)} Case study of Tyrosine kinase activation loop. 
    Ribbon diagram shows the structure of Tyrosine kinase mapped with AlphaFold2 pLDDT score. The activation loop is characterized with low pLDDT scores suggesting flexible conformations;
    \textbf{(B)} Multiple sequence alignment of the activation loop showing this sequence is highly evolutionary conserved. 
    Predictions from ProteinMPNN and \method are shown;
    \textbf{(C)} Preliminary study on refinement ability for \pLMs. Here ESM-1b took as input the predictions of ProteinMPNN;
    \textbf{(D)} Illustration of neural structure-based protein sequence design, and \textbf{(E)} protein language models;
    \textbf{(F)} Overall illustration of \method, where \textbf{the wonderful colored protein structure image is credited to RFDiffusion}~\citep{watson2022RFfold}.}
    \label{fig:main}
    \vspace{-5mm}
\end{figure*}

Albeit deep generative models showing revolutionized capability in this field, we argue that the current neural structure-based protein design approaches are not necessarily at their best in designing more plausible proteins as two major obstacles remain and hinder further progress:
\begin{compactenum}
    \vspace{-5pt}
    \item[(i)] \textbf{Limited experimentally determined protein structure data.}
    For example, the known protein structures in the commonly-used CATH~\citep{orengo1997cath} dataset are multiple orders of magnitude smaller ($<0.1\%$) than the sequence data in the UniRef~\citep{suzek2015uniref} sequence database (\textbf{Fig.~\ref{fig:main}D-E}).
    As structure-based protein design is essentially a conditional sequence learning problem, the protein sequence distribution is crucial yet remains elusive for generally data-hungry generative models due to limited data. 
    Therefore, they fail to holistically explore the protein sequence space and tend to yield sub-optimal sequence predictions for folds. 
    Despite being partly remedied by data-augmented approach~\citep{hsu2022esmif}, additional predicted structure data and trainable model parameters at scale demand compute and storage overheads.

    \item[(ii)] \textbf{Challenge of structurally non-deterministic regions.}
    From a biological perspective, protein structures are sometimes not sufficiently informative, especially for those flexible regions such as loops and exposed surfaces~\citep{towse2012domain}.
    In these regions, residue identities can, hypothetically, be less correlated with the structural context while sequential knowledge is way more useful yet largely neglected.
    We verified this hypothesis and found that existing purely structure-based approaches were prone to produce functionally invalid sequences for these regions (\textbf{Fig.~\ref{fig:main}A-B}).
    \vspace{-3pt}
\end{compactenum}
Therefore, the sequential information should be better utilized for structure-based protein design. 

Inspired by the impressive progress of large language models (LLMs) in natural language processing (NLP)~\citep{devlin2019bert,radford2018gpt,brown2020gpt3}, recent literature in protein research has also demonstrated the emergent evolutionary knowledge of proteins in protein language models~\citep[\pLMs,][]{rives2019esm,lin2022esmfold,hu2022exploring}, learned from the universe of massive protein sequence data.
Such comprehensive and thorough sequential knowledge of \pLMs can help probe functional properties and even predict protein structures from single sequences without the need for explicit evolutionary homologs (\eg, MSAs).
Thus, an exciting research question naturally arises: 
\begin{quote}
\vspace{-6pt}
    Since \pLMs are such strong sequence learners, \emph{can we leverage \pLMs to make better structure-based protein design?}
\vspace{-6pt}
\end{quote}
If so, rather than as protein sequence encoders, \pLMs can possibly be repurposed as sequence generative models (since they are learned to reconstruct corrupted protein sequences), \emph{prompted} by the desired structure to generate sequences, making the most of the acquired sequential evolutionary knowledge.
How to best achieve this goal, however, is non-trivial and remains under-explored (we will discuss our preliminary attempts that uses \pLMs for ``post-editing'' to provide insights that motivate our proposal in \S\ref{sec:prelim}), thus deserves to be comprehensively studied.

In this paper, we show that language models with \textit{structural surgery} are strong protein designers without using abundant training data.
We propose \method, a generic approach to reprogramming sequence-based protein language models (\pLMs) to design protein sequences of a desired fold.
As shown in \textbf{Fig.~\ref{fig:main}F}, we conduct a \emph{structural surgery} on a \pLM (\eg, ESM-1b), where a lightweight \textit{structural adapter }is implanted to endow \pLMs with structural awareness by access to an \textit{arbitrary} additional structure encoder (\eg, ProteinMPNN).
During inference, iterative refinement is performed to optimize the generated protein sequence until convergence when the prediction can no longer be improved.

We highlight our contributions and findings as follows:
\begin{compactitem}
    \vspace{-3pt}
    \item We introduce \method, a generic approach that transforms \pLMs to protein design models via \textit{structural surgery}.
    \method yields preferable protein sequences for desired structures, while being model-agnostic, modularizable, parameter- and data-efficient.
    
    \item Experiments show that \method advances the state-of-the-art methods by a large margin, achieving 55.65\% and 56.76\% sequence recovery on CATH 4.2 and CATH 4.3 for single-chain proteins, and $>$60\% for protein complexes. 
    \method can be also combined with data augmentation~\citep{hsu2022esmif}, where additional large amounts of predicted protein structures by AlphaFold 2~\citep{jumper2021AF2} are leveraged.
    
    \item In particular, we find that \method can accurately handle structurally non-deterministic regions (\eg, functional loops and exposed surfaces) thanks to the learned sequence knowledge from \pLMs, while previous methods typically fail.  We also find that \method can indeed be structurally sensitive, thereby better determining the nuanced sequential specificity of those protein groups of high structural similarity.
    
    \item We also show that \method can synthesize diverse and structurally valid sequences. We further evaluate zero-shot generalizability of \method in designing proteins of unseen categories, including antibodies and \textit{de novo} proteins, and observe superb performance.
\end{compactitem}

We highlight that the goal of this study to propose \method is not to compete but instead to complement current neural structure-based sequence design models. 
We hope that \method can become a powerful, universal, and easy-to-use tool as a ``wrapper'' that helps integrate the advances of both protein sequence learning (\eg, \pLMs) and structure learning (\eg, geometric/graph NNs and protein structure prediction), facilitating future protein research.

\vspace{-2mm}
\section{Preliminaries}
\label{sec:prelim}

\subsection{Structure-based Protein (Sequence) Design}
Structure-based sequence design problem (\emph{a.k.a.}, protein inverse folding) is to find, given a protein backbone structure of interest, an amino acid sequence that will fold to this structure~\citep{dauparas2022proteinmpnn}. 
While physics-based approaches tackle sequence design as an energy minimization problem~\citep{dahiyat1997probing,street1999computational} like the Rosetta~\citep{alford2017rosetta}, recent advances in deep learning (DL) methods have demonstrated great promise in generating plausible amino acid sequences for desired protein structures~\citep{ingraham2019generative,hsu2022esmif}.

\paragraph{Problem Formulation.}
Neural structure-based protein design can be formulated as an end-to-end graph-to-sequence learning problem. 
Formally, a parameterized encoder-decoder neural model $\mathcal{M}_{\theta}$ is tasked with predicting the protein sequence for a protein backbone structure,
\begin{align}
     \mathcal{M}_{\theta}: \mathcal{X} \to \mathcal{S}, \nonumber
\end{align}
where for a protein of length $L$, $\mathcal{S} = \{s_i \in \mathrm{Cat}(20) | 1 \leq i \leq L \}$ is a residue sequence of $20$ types of amino acids, and $\mathcal{X} = \{\bm{x}_i \in \R^{N_\textrm{atoms}\times 3} | 1 \leq i \leq L \}$ denotes the spatial coordinates in 3D space for the residues of the desired protein structure with $N_\textrm{atoms}$ backbone atoms (\eg, $N$, $C_\alpha$ and $C$, with $O$ optionally). 
The learning objective is to find the model parameter $\theta$ that maximizes the conditional log-likelihood $p(\mathcal{S} | \mathcal{X}; \theta)$ given sufficient protein structure-sequence paired data. 
This enables us to design sequences of maximum likelihood, or with sampling algorithms when the diversity and novelty of designs are taken into account.

\noindent\textbf{Overview.} The general workflow of these approaches (\textbf{Fig.~\ref{fig:main}C}) is as follows:
(1) A desired protein backbone structure $\mathcal{X}$ is first represented as a $k$-nearest-neighbor ($k$-NN) graph in 3D space with geometric features attaching to nodes and edges of the graph; 
(2) A graph neural network-based encoder then takes as input the featurized graph and maps it to structural representations; and
(3) Finally, a sequence decoder consumes the encoded structural representations and accordingly predicts a sequence of amino acids $\mathcal{S}$ that is expected to fold into the target protein structure $\mathcal{X}$, in which an autoregressive decomposition $p(\mathcal{S} | \mathcal{X}) = \prod_{i=1}^{L} p(s_i | S_{<i}, \mathcal{X})$ is typically applied. Notably, a recent work \citep{gao2022pifold} found that a fully non-autoregressive factorization~\citep{gu2017non}, \ie, $p(\mathcal{S} | \mathcal{X}) = \prod_{i=1}^{L} p(s_i | \mathcal{X})$ can help achieve a faster and better result.
We will elaborate on the probabilistic models for structure-based protein design (see \S\ref{sec:training}).

\vspace{-2mm}
\subsection{Protein Language Models}
Language models (LMs) trained on large-scale sequence data have shown extraordinary advances and led to a significant paradigm shift in NLP, boosting machines in understanding human languages~\citep[BERT/MLM-style,][]{devlin2019bert} and synthesizing human-like text \citep[GPT/CLM-style,][]{radford2018gpt}.
Analogies between protein sequences and human languages have long been noted~\cite{yang2019machine,ferruz2022controllable}.
As such, it is exciting to expect cutting-edge techniques of language modeling can tackle protein-related problems, especially for generative scenarios such as protein design.

Typically, protein language models (\pLMs, \textbf{Fig.~\ref{fig:main}D}) approximate the protein sequence distribution $p(\mathcal{S})$ via pseudo-likelihood~\citep{salazar2020masked}, wherein $\prod_i p(\mathcal{S}_i | \mathcal{S}_{-i})$ over a partially corrupted sequence (by being randomly masked, replaced or kept up to certain schedules) is maximized.
Although the only training objective is to identify missing amino acids, a high success rate necessitates the model to learn intricate information within its sequential input, \eg, underlying evolutionary correlations and tertiary topology. 

Please refer to Appendix \S\ref{sec:related_work} for more detailed discussions on the related work.



\vspace{-6pt}
\subsection{Proof-of-Concept}
As aforementioned, there are two major challenges for current neural structured-based protein design: (1) from the (conditional) sequence learning perspective, 
\textbi{the lack of abundant protein structure data}, thus a model is hard to properly explore the protein sequence space through data-intensive supervised learning;
and (2) from the biological perspective, \textbi{protein structures are not necessarily always informative}, especially for those flexible regions such as loops and exposed surfaces~\citep{towse2012domain,hsu2022esmif}. 
In these cases, residue identities are less correlated to the spatially associated tertiary structure while sequential evolutionary knowledge can be more decisive.
For example, as shown in \textbf{Fig.~\ref{fig:main}A-B}, the activation loop in Tyrosine kinase has multiple conformations and is not well spatially constrained. However, it is known to play an important function in regulating the activity of the protein. 
When pure structure-based modes, \eg ProteinMPNN~\citep{dauparas2022proteinmpnn}, is used to design this functional loop, it was prone to produce functionally invalid repeated sequences.


Based on the above observations, we conjecture that protein language models (\pLMs) can have promise for circumventing these two problems since they have learned from large-scale protein sequence data, thereby protein sequential evolutionary knowledge is supposedly acquired.
To verify our hypothesis, we started with a quick sanity check by studying whether a trained \pLM has the capability to refine predicted sequences generated by currently advanced strong structure-based approaches.
Specifically, we inspected the pretrained ESM-1b~\citep{rives2019esm} for \pLM together with the ProteinMPNN~\cite{dauparas2022proteinmpnn} for the protein design model, both are popular and openly accessible.
We tested on CATH 4.2 test split, where sequence predictions of the ProteinMPNN were fed as input into the ESM-1b.

As shown in \textbf{Fig.~\ref{fig:main}C}, without any training, an immediate accuracy improvement is achieved.
This promising observation boils down to two properties of \pLMs: (1) acquired sequential evolutionary knowledge, and (2) learning to denoise from corrupted protein sequences.
This sheds light on the use of \pLMs to boost protein sequence design.
Note that in this naive \textit{post-editing} manner, \pLMs cannot access the structure directly, hence the potential of exploiting structural information to steer \pLMs for structure-based sequence design is not fully utilized. 
In the next section, we will take a further step to devise a generic framework that makes sequence-level \pLMs and structures best integrated by the proposed \emph{structural surgery}.

\vspace{-2mm}
\section{Reprogramming \pLMs for Structure-based Protein Design with Structure Surgery}
\label{sec:method}

\paragraph{Overview.}
We proposed a generic framework named \method that reprograms \pLMs to yield high-quality protein sequences from desired backbone structures.
A general illustration is shown in \textbf{Fig.~\ref{fig:main}F}. 
To endow pLMs with structural awareness, we introduce a lightweight \textit{structural adapter}~\citep{houlsby2019parameter} into \pLMs (\eg., ESM-1b) such that the emergent ability of acquiring protein geometric information from an arbitrary structure encoder is unlocked.
During inference, iterative refinement is performed to optimize the generated protein sequence until the prediction could no longer be improved.

For brevity, we will take ProteinMPNN and ESM-1b as a running example hereafter.
Note that, however, the proposed framework is model-agnostic, regardless of the choices for any structure encoders and \pLMs.
It can accommodate any other strong structure-based design models, \eg, PiFold~\citep{gao2022pifold}, and ESM-2 series as \pLM alternatives.
This also implies that \method can be highly modularizable, especially when pretrained parameters of the structure encoder are available.
These two merits help \method attain further performance bonus (see \$\ref{sec:experiment}).



\vspace{-5pt}
\subsection{Training}
\label{sec:training}
To better accommodate \pLMs that are tasked with masked language modeling ~\citep[MLM,][]{devlin2019bert} as the learning objective, our framework is accordingly established in a denoising auto-encoding fashion. 
More concretely, we use conditional masked language modeling~\citep[CMLM,][]{ghazvininejad2019mask}, which is proven to be very useful for conditional sequence generation in NLP~\citep{gu2017non}, as the learning scheme.
CMLM shares similarities with the BERT-style objective of sequence representation learning, but is more suitable for the generative purpose. 
Formally, given backbone structure $\mathcal{X}$ and sequence $\mathcal{S} = S_{\text{masked}} \cup S_{\text{obs}}$, CMLM requires the model to predict a set of target amino acids $S_{\text{masked}}$, which are randomly masked, from the remaining observed residues
\begin{align}
    p(S_{\text{masked}} | S_{\text{obs}}, \mathcal{X};\theta) = \prod_{s_i \in S_{\text{masked}}}p(s_i | S_{\text{obs}}, \mathcal{X}; \theta), \nonumber
\end{align}
Here a conditional independence assumption over identities of target residues $s_i \in S_{\text{masked}}$ is made, given $\mathcal{X}$ and $S_{\text{obs}}$.

We remark that such a conditional independence assumption is almost true for structure-based protein design from the viewpoint of probabilistic graphical models (PGMs), wherein graphically represented protein structure implies that each amino acid is primarily dependent on its spatially nearest neighbors rather than considerably distant ones.
In our preliminary experiments, we found that existing strong models (\eg, ProteinMPNN) trained with CMLM consistently outperformed their original autoregressive formulation, where only ``left'' contexts rather than all neighbors get considered. 
This indicates that such an assumption can effectively exploit the underlying structural information, thereby better formulating structure-based protein design.

\begin{table*}[t]
   \centering
   \small
   \vspace{-2.5mm}
   \caption{{\sl Performance comparison between \method and different baseline approaches on CATH 4.2 and CATH 4.3 datasets.} {\method's results are obtained by 5-cycle iterative refinement with \texttt{argmax} operator (\ie, no sampling).}
   $\dagger$: benchmarked results are quoted from \citet{gao2022pifold}.
   $\ddagger$: ``Single-chain'' in \citet{hsu2022esmif} is defined differently. ESM-1b 650M is used as the default \pLM.}
   \vspace{1.5pt}
   \label{tab:results_cath}

   \resizebox{\linewidth}{!}{%
   \begin{tabular}{llrcccccc}
   \toprule
   & \multirow{2}{*}{\bf Models} 
   & \multirow{2}{*}{\bf Trainable/Total} 
   & \multicolumn{3}{c}{\bf Perplexity ($\downarrow$)} 
   & \multicolumn{3}{c}{\bf Median Recovery (\%, $\uparrow$)} \\
   \cmidrule[0.3pt](lr){4-6} \cmidrule[0.3pt](lr){7-9}
   & & \bf Params. & Short   & Single-chain  & All  & Short   & Single-chain   & All \\
   \midrule

   \multirowcell{10}{\rotatebox[origin=c]{90}{CATH 4.2}}
   & $^\dagger$Structured Transformer~\smallcitep{ingraham2019generative}                     & 1.56M/1.56M  & 8.39  & 8.83 & 6.63&   28.14   &28.46& 35.82 \\
   & $^\dagger$GVP~\smallcitep{jing2020gvp}                                       & 1.0M/1.0M  & 7.23  &   7.84 &  5.36 &   30.60 &28.95 &  39.47  \\
   & $^\dagger$ProteinMPNN~\smallcitep{dauparas2022proteinmpnn}                   & 1.9M/1.9M  & 6.21  &   6.68  &  4.61 &  36.35  & 34.43  &  45.96   \\
   & PiFold~\smallcitep{gao2022pifold} \                                & 6.6M/6.6M  & {6.04}  &   {6.31}  &  {4.55} &   {39.84} &{38.53} &  {51.66} \\ 
  \cmidrule[0.5pt](lr){2-9}
  & ProteinMPNN + CMLM [ProtMPNN-CMLM]                                                 & 1.9M/1.9M  &7.16 &7.25  & 5.03    &35.42 &35.71&  48.62  \\ 
  & \chl \method (ProtMPNN-CMLM)                                          & \chl 6.9M/659M (\textbf{1.0\%})  &\chl 7.02 & \chl 6.67 & \chl 4.63   & \chl 35.71  & \chl 38.61 & \chl 53.26 \\ 
  & \chl \method (pretrained ProtMPNN-CMLM: \textit{fine-tune})           & \chl 6.9M/659M (\textbf{1.0\%})  & \chl 7.33 & \chl 6.83 & \chl 4.63 & \chl \textbf{36.47} & \chl \textbf{40.91} & \chl \textbf{54.62} \\
  & \chl \method (pretrained ProtMPNN-CMLM: \textit{freeze})              & \chl 5.0M/659M (\textbf{0.7\%})  & \chl \textbf{7.01} & \chl \textbf{6.58} & \chl \textbf{4.41} & \chl 35.19 & \chl 40.00 & \chl 54.41 \\
  \cmidrule[0.2pt](lr){2-9}
  & PiFold (reimpl.)                                                    & 6.6M/6.6M  & 6.94 &6.82  &4.69  &36.67 &38.21&  50.22   \\ 
  & \chl \method (PiFold)                                               & \chl 11.9M/664M (\textbf{1.7\%})  & \chl \textbf{6.77} & \chl \textbf{6.46}   & \chl \textbf{4.52}   & \chl \textbf{37.88}   &\chl \textbf{42.47}   & \chl \textbf{55.65} \\ 
  \midrule

  \multirowcell{10}{\rotatebox[origin=c]{90}{CATH 4.3}}
  & GVP-large~\citep{hsu2022esmif}                                      & 21M/21M  &  7.68  & $^\ddagger$6.12~~ &  6.17 &   32.60 & $^\ddagger$39.40~~ &   39.20   \\
  & GVP-Transformer~\citep{hsu2022esmif}                                 & 142M/142M  &   8.18 & $^\ddagger$6.33~~ &  6.44 &   31.30 & $^\ddagger$38.50~~ &  38.30 \\
  & ~~~~+1.2M AF2 predicted data~\smallcitep{hsu2022esmif}              & 142M/142M  &   6.05 & $^\ddagger$4.00~~ &  4.01 &   38.10 & $^\ddagger$51.50~~ &  51.60 \\
  & \chl \method (pretrained GVP-TransEncoder: \textit{freeze})          & \chl 6.3M/800M (\textbf{0.8\%})  & \chl 6.61 & \chl 7.64 & \chl \textbf{3.92} & \chl 42.31 & \chl 38.64 & \chl \bf 56.49 \\
  \cmidrule[0.5pt](lr){2-9}
  & ProteinMPNN + CMLM                                                  & 1.9M/1.9M  & 6.31 &6.32  &4.85     & 40.30 & 39.02 &  48.25  \\ 
  & \chl \method (ProtMPNN-CMLM)                                          & \chl 6.9M/659M (\textbf{1.0\%})  & \chl 5.96 &\chl 5.85 &\chl 4.48   & \chl 44.12  & \chl 44.05 &\chl 54.05 \\ 
  & \chl \method (pretrained ProtMPNN-CMLM: \textit{fine-tune})           & \chl 6.9M/659M (\textbf{1.0\%})  & \chl 5.94 & \chl 5.80 & \chl 4.20 & \chl 45.65 & \chl \textbf{46.15} & \chl 55.92 \\
  & \chl \method (pretrained ProtMPNN-CMLM: \textit{freeze})              & \chl 5.0M/659M (\textbf{0.7\%})  & \chl \textbf{5.88} & \chl \textbf{5.66} & \chl \textbf{4.19} & \chl \textbf{45.71} & \chl \textbf{46.15} & \chl \textbf{56.38} \\
  \cmidrule[0.2pt](lr){2-9}
  & PiFold (reimpl.)                                                    & 6.6M/6.6M  & 5.88& 5.55 & 4.47  & 42.86 &43.69& 50.68    \\ 
  & \chl \method (PiFold)                                               & \chl 11.9M/664M (\textbf{1.7\%})  &\chl \textbf{5.66} &\chl \textbf{5.52}  &\chl 4.01   & \chl \textbf{46.84}  &\chl \textbf{48.63}  & \chl \textbf{56.63} \\ 
  
   \bottomrule
   \end{tabular}
   }
   \vspace{-8pt}
\end{table*}

\vspace{-5pt}
\subsection{Inference with Iterative Refinement}
\label{sec:inference}

To predict protein sequences from a given structure, our goal is to sample sequences $\hat{\mathcal{S}}$ that have high likelihoods \wrt $p(\mathcal{S} | \mathcal{X})$. 
Particularly, $\hat{\mathcal{S}}$ with the maximum likelihood can be obtained via greedy deterministic decoding: $\hat{\mathcal{S}} = \argmax_{\mathcal{S}} p(\mathcal{S} | \mathcal{X}) = \argmax_{\mathcal{S}=\{s_i | 1 \leq i \leq L\}} \prod_i p(s_i | \mathcal{X})$. 
Notably, \method is trained to reconstruct a protein native sequence from its corrupted version, which enables it to iteratively refine the predicted sequence in a coarse-to-fine manner towards a better one~\cite{savinov2021step}.

Concretely, to sample in such an iterative refinement manner,  we follow the structure of the Markov process and sample sequentially $ {\mathcal{S}}^{(t)} \sim p(\mathcal{S}^{(t)} | \mathcal{S}^{(t-1)}, \mathcal{X})$ by recycling the \pLM-based decoder for some ﬁxed number of steps $T$, starting from an initial sequence ${\mathcal{S}}^{(0)}$.
Here the initial sequence ${\mathcal{S}}^{(0)}$ can be drawn from a weaker proposal distribution parameterized by a simple linear projection to amino acid vocabulary from the features of the structure encoder\footnote{A ``full-mask'' initialization also works well but considerably sub-optimal.}. 
In our primary case where ProteinMPNN~\citep{dauparas2022proteinmpnn} serves as the structure encoder, we let this linear projector be its original output head, hereby ${\mathcal{S}}^{(0)}$ can be regarded as the output of the ProteinMPNN.

We can tune the number of steps $T$ for a good accuracy-efficiency trade-off.
Larger $T$ usually leads to better prediction with high latency, whereas one-shot parallel generation, as a special case, can be achieved by setting $T = 1$ when efficiency is prioritized.



\paragraph{Temperature-based Sampling.}
To control the diversity and speed of convergence, we consider a modified function of the categorical distribution we want to sample from such that $\log p(\mathcal{S} | \cdot; \tau) \propto \frac{1}{\tau} \log p(\mathcal{S} | \cdot)$, where $\tau$ is the temperature.

  


  
  

  


\section{Experiments}
\label{sec:experiment}

In this section, we evaluate \method on a variety of benchmarks for fixed backbone protein sequence design (including single-chain proteins and multi-chain protein complexes).
We start with evaluations on designing single-chain proteins using standard CATH 4.2 and 4.3~\cite{orengo1997cath} benchmarks as in ~\citep{ingraham2019generative,hsu2022esmif}. 
We then go for assessing capabilities of designing multi-chain complexes using the dataset curated and used in \citet{dauparas2022proteinmpnn}. 
We instantiate the structure encoders in \method using the same hyperparameters as the corresponding backbone structure models, and the dimensions of the structural adapter are thereby determined.
Moreover, comprehensive and thorough analyses on diverse settings of our interest are conducted thereafter, including sequence accuracy \vs diversity, effects of scaling \wrt data and model size, structural validity, and generalizations to other proteins (\eg, antibodies and \textit{de novo} proteins).
Finally, systematic bio-informatics assessments are presented to help dive deep into the intrinsic properties of \method.

Please refer to Appendix for more detailed experimental settings and additional results of all the evaluations.

\subsection{Benchmarking Fixed Backbone Protein Design}

\subsubsection{Single-chain Protein Design}

\textbf{Tab.~\ref{tab:results_cath}} shows the results of \method in comparison to the recent strong baselines on the CATH~\cite{orengo1997cath} benchmark, including the current strongest ones. 
We highlight our primary findings as follows:

\noindent\textbf{(1) \method is more data-efficient and advances state-of-the-arts methods with a large margin without any additional data.}
On the more commonly-used CATH 4.2 benchmark, we observe that improving protein featurizers capability structure encoders, from the vanilla message-passing (graph) neural networks~\citep{ingraham2019generative} to more complicated ones~\citep{jing2020gvp,dauparas2022proteinmpnn,gao2022pifold}, indeed facilitate performance to some extents, but are limited to the under-representative protein sequence distribution due to the data shortage of experimentally determined CATH datasets, which is expected to be mitigated by \pLMs.
Our motivation has primarily been verified through our illustrative ProteinMPNN + ESM-1b setting, wherein by taking advantage of massively pretrained \pLMs, the proposed \method improves ProteinMPNN + CMLM by 4.4\% on CATH 4.2 (48.62\%$\to$52.99\%) and 5.8\% on CATH 4.3 (48.25\%$\to$54.05\%), setting the new state-of-the-arts without using any augmented data.

\noindent\textbf{(2) \method can be modularizable and further benefit from pretrained structure models.}
Instead of learning structure encoders from scratch, \method can leverage pretrained structure models as encoders, which can be fine-tuned together with the structural adapter or kept frozen. 
As little difference in sequence recovery, we suggest that freezing is the best practice.
In this case, only a tiny proportion of parameters (\ie, those of the structural adapter) are trainable, and \method quickly converges with better results in a negligible overhead of 10 epochs.
Notably, we also study \method with encoder from GVP-Transformer~\cite{hsu2022esmif}, which was built on 1.2M AF2 predicted data, giving rise to a $\sim$5\% further improvement.

\noindent\textbf{(3) The more advanced the structure encoder becomes, the stronger our model-agnostic \method performs.}
Since our aim is to devise \method as a general-purpose framework that can make the most of the progress of protein structure models, we study whether \method can get boosted as if meets stronger structure encoders. 
We built a variant of \method upon PiFold~\citep{gao2022pifold}, a most recent and performant structure-based design model. 
As we can tell from \textbf{Tab.~\ref{tab:results_cath}}, \method improves PiFold by at least 5.43\% in terms of sequence recovery, yielding impressive 55.65\% and 56.63\% on CATH 4.2 and 4.3 datasets.
These results demonstrate that \method is a general-purpose approach for structure-based sequence design, which is compatibility-friendly and hence fully leverages the advances of protein structure learners. 

Combining \textbf{(2)} and \textbf{(3)}, \method helps structure-based protein design make more gains with less pain, thanks to the breakthrough progress and the open-source efforts of the whole community, standing on the shoulders of giants.


\subsubsection{Multi-chain Protein Complex Design}

\begin{table}[t]
\centering
\small
\vspace{-1.5mm}
\setlength{\tabcolsep}{1pt}
\caption{{\sl Performance on multi-chain protein complex dataset (in median recovery).} Results of the original ProteinMPNN and GVP-Transformer were obtained using publicly available checkpoints. }
\label{tab:results_complex}
\vspace{1.5pt}
\resizebox{\linewidth}{!}{%
\begin{tabular}{lr} 
\toprule
\bf Models                        & \bf Rec. ($\uparrow$)           \\
\midrule
ProteinMPNN~\smallcitep{dauparas2022proteinmpnn}                      & 50.00      \\
\midrule
ProteinMPNN + CMLM [ProtMPNN-CMLM]                                  & 54.39     \\
\chl \method (ProtMPNN-CMLM + ESM-1b 650M)                                & \chl 59.10   \\
\chl \method (pretrained ProtMPNN-CMLM: \textit{freeze})                                & \chl \bf 59.43   \\
\chl \method (pretrained ProtMPNN-CMLM: \textit{fine-tune})                                & \chl \bf 59.43   \\

\midrule
\chl \method (ProtMPNN-CMLM + ESM-2 650M)                                & \chl 59.81   \\
\chl \method (ProtMPNN-CMLM + ESM-2 3B)                                & \chl \bf 61.49   \\
\midrule
\chl \method (pretrained GVP-TransEncoder.:\textit{freeze} + ESM-1b 650M)                                & \chl \bf 62.16   \\
\bottomrule
\end{tabular}
}
\vspace{-3mm}
\end{table}

A protein functions only when it docks, combines, and interacts with other macro- and macro-molecules, composing multi-chain protein complexes. 
As such, studying protein sequence design for multi-chain assemble structures is crucial for drug design.
This incentivizes us to evaluate whether \method can better handle protein complex design.


We use the multi-chain complex dataset curated by \citet{dauparas2022proteinmpnn} and use the same training settings 
as in our single-chain scenario.
We summarize our results regarding protein complex design in \textbf{Tab.~\ref{tab:results_complex}}. 
We first notice that CMLM~\citep{ghazvininejad2019mask} can better formulate and train ProteinMPNN than the original autoregressive version with teacher-forcing.
Upon a more competent system of ProteinMPNN + CMLM, \method yields a near 60\% sequence recovery of multi-chain protein assembles. 
When further integrated with \pLMs at scale (\ie, ESM-2) or better structure encoder (\ie, GVP-TransformerEncoder), it can even achieve more impressive scores of 61.49\% and 62.16\%, respectively.
These results show that \method can not only design single-chain proteins, which are mostly studied in previous works but also be used for designing multi-chain protein complexes. 
This makes \method more general-purpose in terms of the categories and scenarios where it can be deployed, and opens opportunities to use it for designing specific protein complexes (see \S\ref{sec:zero_shot}), such as antigen-antibody or protein-ligand assemblies.

\begin{figure}[t]
    \centering
    \!\!\!\includegraphics[width=0.9\linewidth]{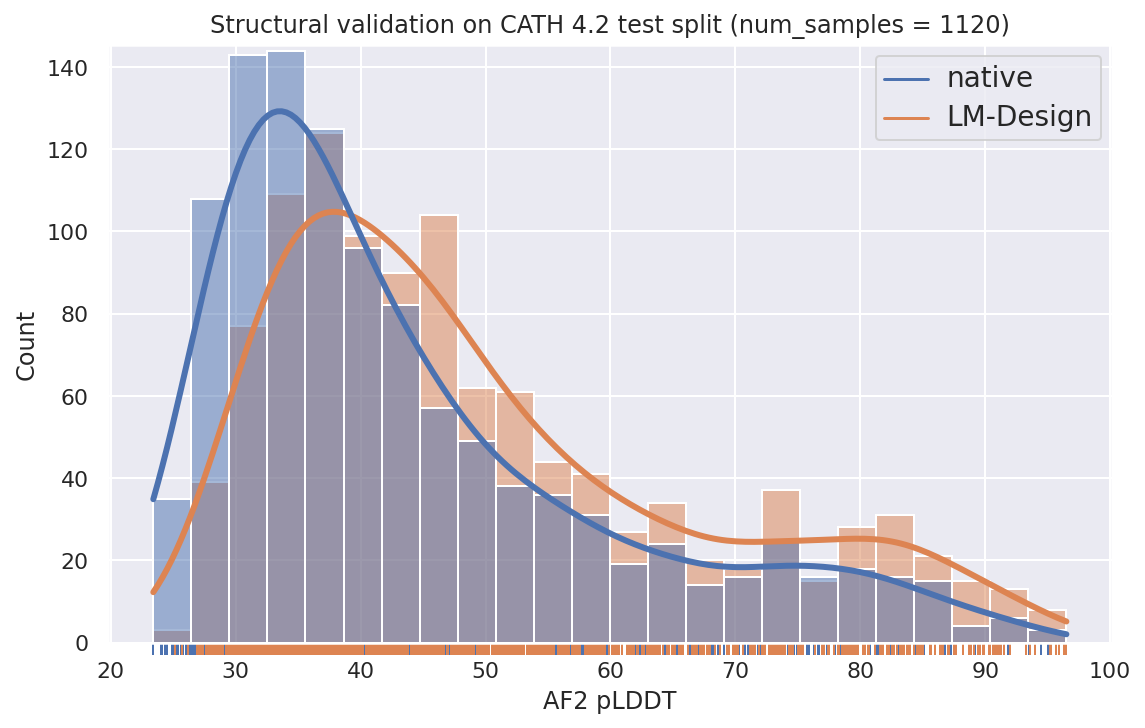}
    \vspace{-15pt}
    \caption{Structural validation measured by the distribution of AF2 pLDDT ($\to$) for both native and \method (ProteinMPNN) redesigned sequences. }
    \label{fig:structural_validation}
    \vspace{-2mm}
\end{figure}

\subsection{Analyses: Diving Deep into \method}

\subsubsection{Structural Validity using AlphaFold 2}
Given that experimental assessment is not available, we leverage the most famous \emph{in silico} structure prediction protocol, \ie AlphaFold 2~\citep[AF2,][]{jumper2021AF2}, to evaluate the structural validity of our designs. 
Here we use the pLDDT score of AF2 as the evaluation metric and follow the evaluation configurations as in \citet{dauparas2022proteinmpnn}, where AF2 takes as input only single sequences (native sequences and our designs) while no multiple sequence alignments (MSAs) are provided.
We redesign all 1120 proteins in CATH 4.2 test split.
As shown in \textbf{Fig.~\ref{fig:structural_validation}}, \textbi{\method's redesigns are predicted, by AF2, to adopt the given backbone structures more confidently than the native sequences}, implying higher structural stability of our redesigns.
We postulate this is because, in this challenging setting where no co-evolutionary information of homologous sequences is exposed to AF2, \method exploits the full potential of sequential co-evolutionary knowledge that \pLMs learns from massive protein sequences.

\begin{figure}[t]
    \centering
    \!\!\!\!\!\!\!\includegraphics[width=0.9\linewidth]{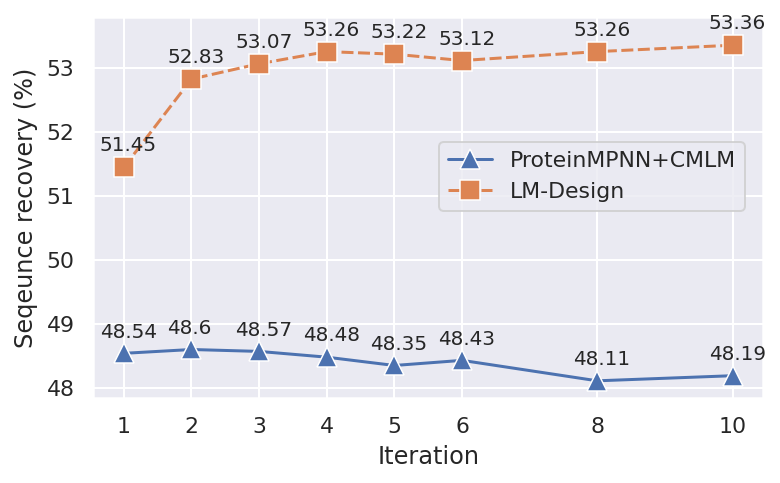}
    \vspace{-15pt}
    \caption{{\sl Effects of iterative refinement}: sequence recovery \wrt decoding iterations.}
    \label{fig:iterative}
    \vspace{-1mm}
\end{figure}

\begin{figure}[t]
    \centering
    \vspace{-8pt}
    \!\!\!\!\!\!\!\includegraphics[width=0.9\linewidth]{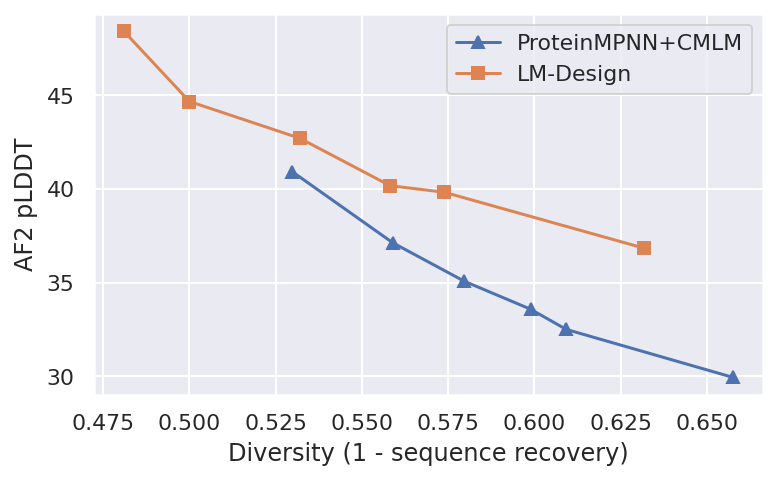}
    \vspace{-15pt}
    \caption{{\sl Pareto optimal points that represent the trade-off between accuracy (in AF2 pLDDT) and diversity}, showing accurate and diverse predictions \method can make. }
    \label{fig:acc_diversity}
    \vspace{-2mm}
\end{figure}

\subsubsection{Studies on Inference}

\paragraph{Iterative refinement gives rise to accurate sequence design.}
Since \method is trained to denoise, we can exploit iterative decoding to progressively refine its predictions towards a better one (\S\ref{sec:inference}). 
As shown in \textbf{Fig.~\ref{fig:iterative}}, we find that even without iterative refinement \method performs sufficiently well, while recycling the \pLM-based decoder for \method yields 1$\sim$2\% gains.
This shows that iterative refinement is an effective strategy for sequence design if models are set up under a denoising learning scheme.
Significant further improvement eliminates if iterating beyond 6 rounds, resulting in acceptable sampling efficiency.

\paragraph{Accuracy \vs diversity.}
While recent protein sequence design approaches have focused on maximizing native sequence recovery (and we also use it as our primary metric in experiments above), this is not necessarily optimal for actual protein design applications, for which novelty also matters~\citep{dauparas2022proteinmpnn}.
To this end, we manipulated the temperatures ($\tau \in [0.1, 0.5, 1.0, 1.2, 1.5]$) to control the diversity of sampled sequences that are dissimilar to the native ones at different levels (as in \S\ref{sec:inference}), and evaluate design accuracy (in AF2 pLDDT) as a function of diversity.
\textbf{Fig.~\ref{fig:acc_diversity}} shows that \textbi{\method yields diverse yet more accurate designs} over ProteinMPNN, manifesting the potential of practical values of \method in real-world scenarios.

\begin{figure}[t]
    \centering
    \vspace{-2pt}
    \!\!\!\!\includegraphics[width=0.9\linewidth]{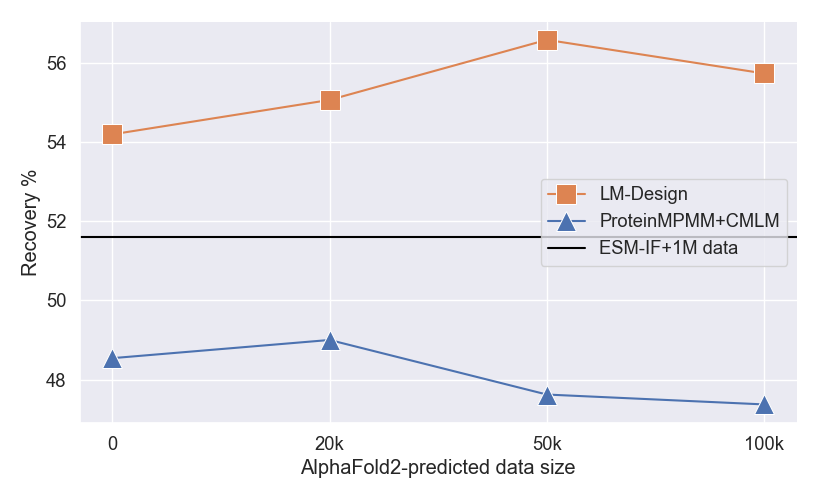}
    \vspace{-15pt}
    \caption{{\sl Performance \wrt scales of augmented data.} Predicted data are produced by AlphaFold 2 from SWISS-PROT dataset.}
    \label{fig:dataaugmentation}
    \vspace{-2mm}
\end{figure}

\subsubsection{Does Scaling Help? Ablations on Scales of Structure Data and \pLMs}
\paragraph{\method works well with data augmentation~\cite{hsu2022esmif} via incorporating predicted structures from AlphaFold 2.}
We perform different scales of data augmentation, the details of data processing are described in Appendix~\ref{appendix:B.5}.  As the results in \textbf{Fig.~\ref{fig:dataaugmentation}} show that both of the methods obtain better results with 20k data augmentation. 
While ProteinMPNN+CMLM drops at 50k, \method keeps increasing and finally drops at 100k. That is because \method has 6.9M parameters while ProteinMPNN+CMLM only has 1.6M parameters. 
The same conclusion has also been mentioned in ESM-IF~\citep{hsu2022esmif}, where models of less capacity cannot benefit from the increment of data.

\begin{figure}[t]
    \centering
    \includegraphics[width=0.99\linewidth]{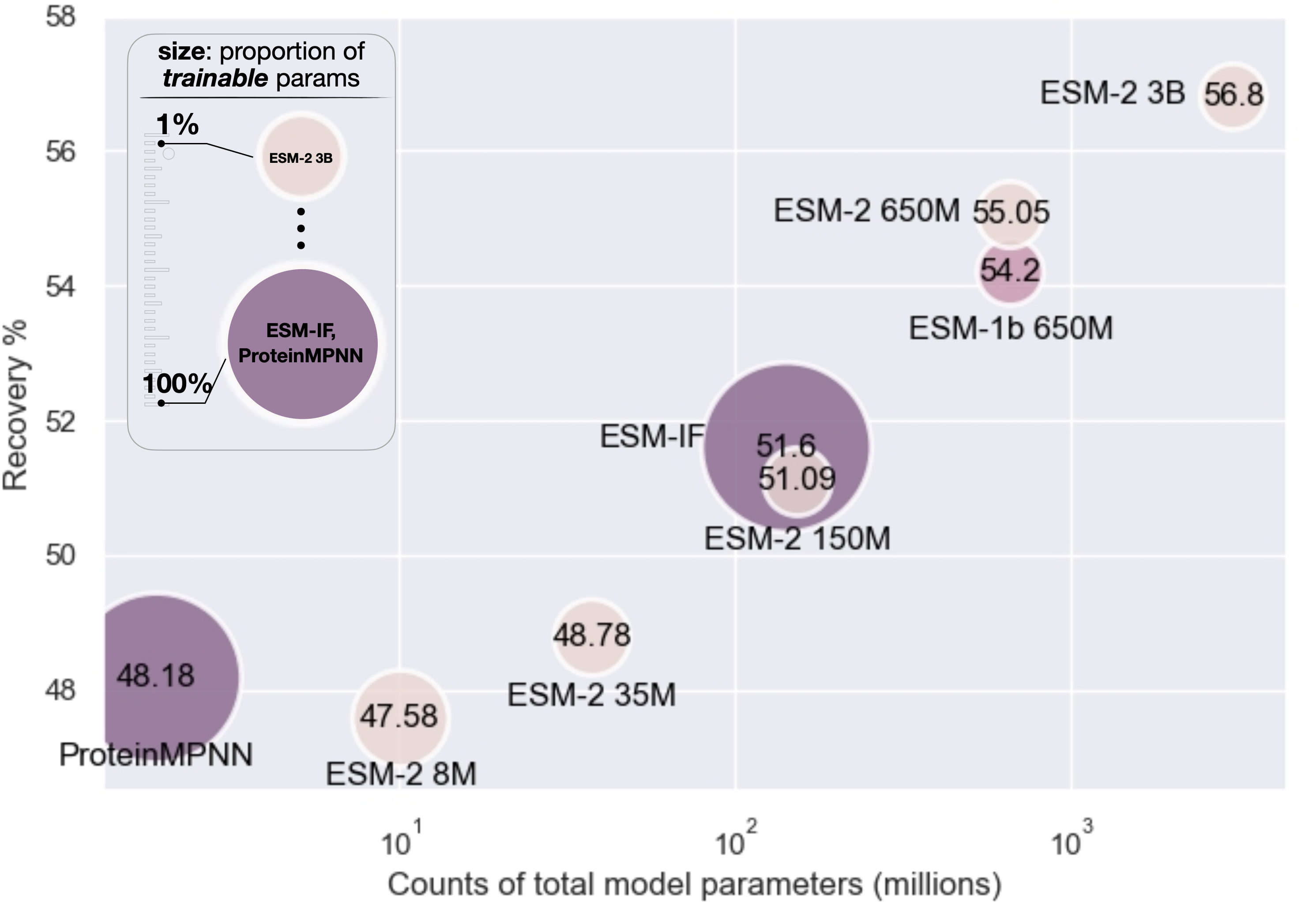}
    \vspace{-10pt}
    \caption{{\sl Performance \wrt model scales of \pLMs using ESM-2 series on CATH 4.3.}
    \textbf{X-axis}: counts of total model parameters; \textbf{Y-axis}: sequence recovery ($\uparrow$); and the \textbf{bubble size}: proportions of trainable parameters over total model parameters ($\downarrow$).}
    \label{fig:model_scales}
    \vspace{-1pt}
\end{figure}

\paragraph{\method is scalable yet parameter-efficient: scaling law \wrt model sizes of \pLMs using ESM-2 series~\citep{lin2022esmfold}.}
To study the impact of the scale of \pLMs, we switch \method's decoder from ESM-1b (145M) to ESM-2 series, with parameters (params) ranging from 8M to 3B.
As shown in \textbf{Fig.~\ref{fig:model_scales}}, the performance of \method increase as model scaling, namely the larger the better, with a clear (log) scaling law exhibits. 
Such a scaling law agrees with large language models (LLMs) in general, which was initially observed and well-known in  NLP~\citep{kaplan2020scaling}.
In addition, in contrast to existing strong systems (\eg, ESM-IF~\cite{hsu2022esmif} requires expensive training of all hundred millions of params in total), \method is way more parameter-efficient only needing $<1\%$ trainable params \wrt total params of the corresponding \pLMs.
In the extreme case, the largest ESM-2 3B-based variant has \textbf{22M} trainable out of \textbf{3B} total params (\textbf{0.07\%}) and achieves the highest accuracy of 56.8\%.
We suggest that this strong connection between protein and natural languages on LLMs gives rise to exciting potentials to empower protein research with cutting-edge advances in general AI.

\begin{figure}[t]
    \vspace{-1.5mm}
    \centering
    \includegraphics[width=\linewidth]{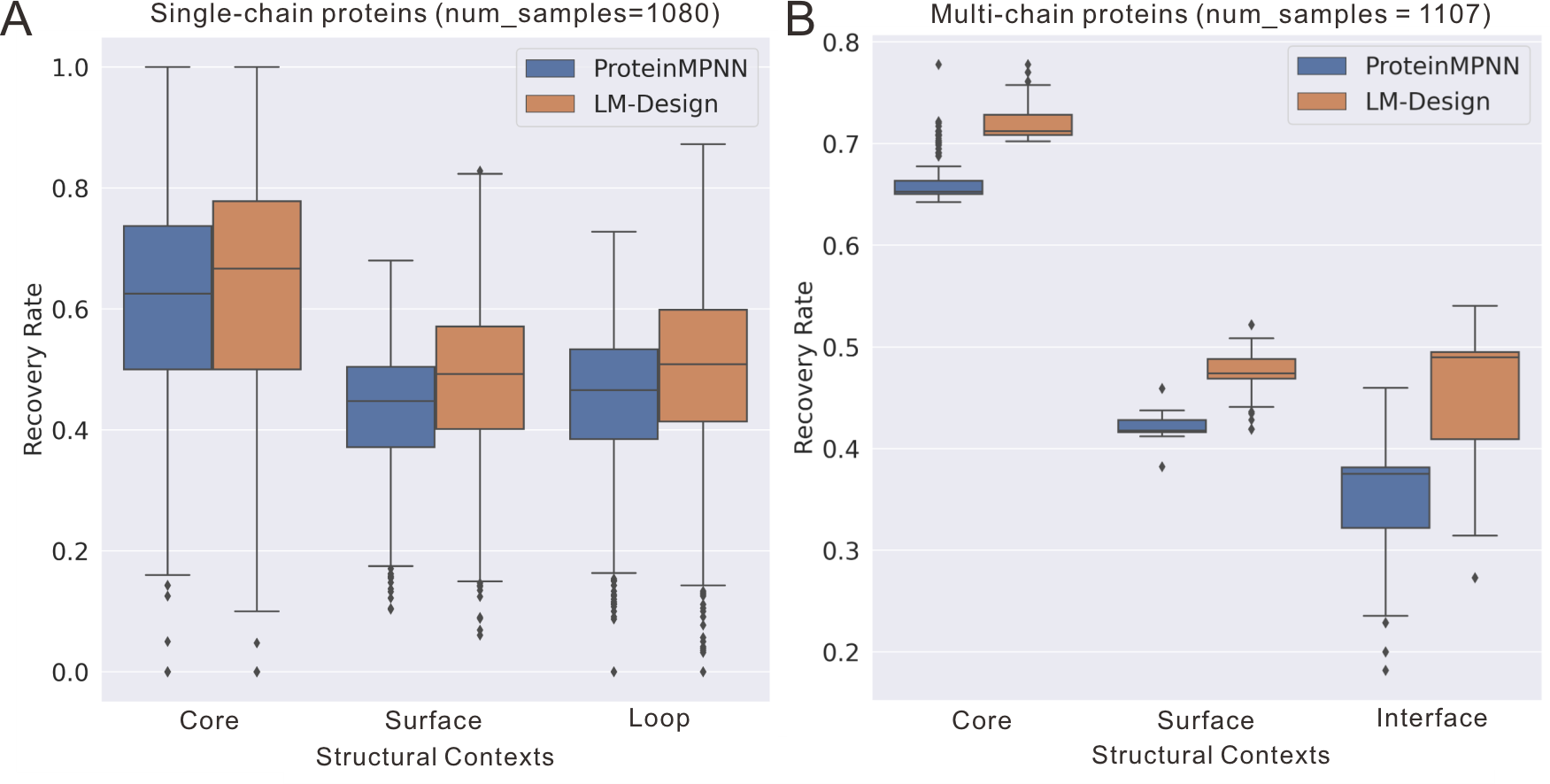}
    \vspace{-21pt}
    \caption{{\sl Comparison of sequence recovery \wrt structural contexts regarding solvent accessible surface area and interaction interface}, on (A) CATH 4.2 single-chain proteins and (B) multi-chain protein complexes.  }
    \label{fig:substructure}
    \vspace{-2mm}
\end{figure}

\subsubsection{How \method Improves Protein Design? Studies on Protein Structures}
\vspace{1pt}
\paragraph{\method effectively exploits the potential of both structural and sequential capabilities.}
To further understand the action mechanism of \method, 
we dissect its performance based on distinct structural contexts either with high structural constraints or low constraints. As shown in \textbf{Fig.~\ref{fig:substructure}A}, for single-chain proteins in the CATH dataset, structured-based ProteinMPNN shows high sequence recovery rates on structurally constrained residues in the folding core, while low recovery rates on structurally less-constrained residues on surface area and loops. Interestingly, \method can effectively enhance the sequence recovery rates on structurally-constrained and less-constrained residues. Similar observations can be found for multi-chain complex proteins as shown in \textbf{Fig.~\ref{fig:substructure}B}. Although ProteinMPNN achieves high sequence recovery rates on folding core residues, it shows compromised performance on residues in the binding interface and exposed regions. \method can generally improve sequence recovery rates in different structural contexts.

\begin{figure}[t]
    \centering
    \includegraphics[width=0.98\linewidth]{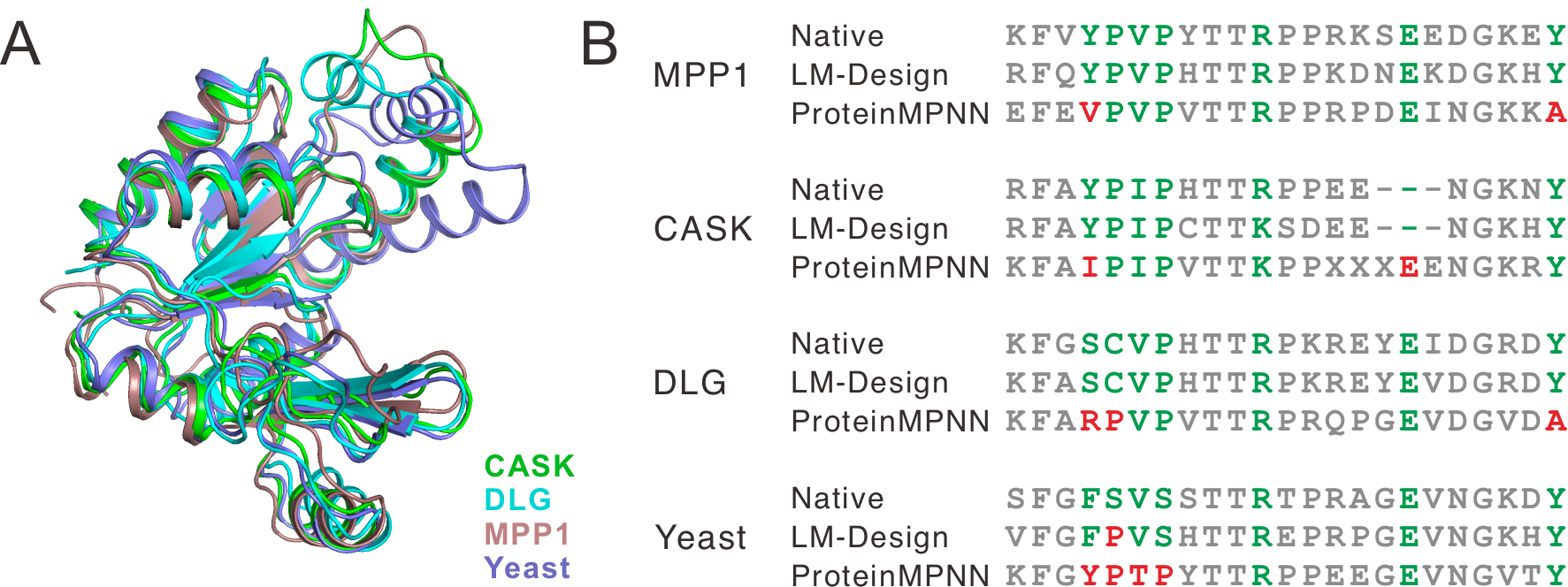}
    \vspace{-8pt}
    \caption{{\sl Case study to evaluate the structure sensitivity of the methods.} (A) Structure alignments show highly similar structures of the four proteins. (B) Multiple sequence alignment. These distinct residues mediating specific biological functions are highlighted in \textcolor{emerald}{\textbf{green}}.
    The sequence predicted by \method and ProteinMPNN were aligned to native sequences. 
    The incorrect predictions are highlighted in \textcolor{red}{\textbf{red}}.}
    \label{fig:sensitivity}
    \vspace{-3mm}
\end{figure}

\paragraph{\method is structurally sensitive.}
It is interesting to know whether the sequence-based protein language model \method is sensitive to structure inputs. To evaluate this problem, we followed a previous biological study~\citep{zhu2011guanylate}, collected four proteins sharing similar structures but having distinct sequences with specific functions. We tested \method's performance using this case for designing specific functional sequences. As shown in \textbf{Fig.~\ref{fig:sensitivity}}, \method can nicely predict functional specific sequences for each of the proteins, suggesting \method is highly sensitive to structure variations.

\subsection{Zero-shot Generalization to Other Proteins}
\label{sec:zero_shot}
\vspace{1pt}

\paragraph{TS50 and TS500 datasets.}
TS50 and TS500 are commonly used independent test set to assess model generalization for proteins introduced by \citet{Li2014TS50}. 
We test \method trained on CATH 4.2 and 4.3 respectively. 
As shown in \textbf{Tab.~\ref{tab:results_TS50TS500}}, we can first find that models trained on CATH 4.3 seem to generalize better to these two datasets, and \method consistently achieves the best.
As for TS500, in particular, we observe that \method improves PiFold (58.82\% $\to$67.78\%) and outperforms previous methods by a large margin.
Notably, non-\pLM-enpowered models (\eg, Pifold, though we re-implemented with comparable performance), sometimes assign way too higher probabilities to the incorrect amino acids, especially for the bad cases we observed in TS500, giving rise to enormous perplexity (marked as \texttt{n/a}).
With the help of \pLMs, \method (PiFold) goes back to a normal scale of perplexity, indicating the capabilities of \pLMs again.


\begin{table}[t]
   \centering
   \small
   \setlength{\tabcolsep}{2pt}
   \vspace{-2.2mm}
   \caption{{\sl Performance comparison on the TS50 and TS500 datasets.} We follow previous literature to mainly report the results using models trained on CATH 4.2. Results from models trained on CATH 4.3 are also provided  (numbers in brackets). DenseCPD~\smallcitep{qi2020densecpd} is a CNN-based (not GNN-based) approach.} 
   \vspace{1.5pt}
   \label{tab:results_TS50TS500}
   \resizebox{1\linewidth}{!}{%
   \begin{tabular}{lcccccc}
   \toprule
    \multirow{2}{*}{\bf Models} & 
    \multicolumn{3}{c}{\bf TS50 }  &
    \multicolumn{3}{c}{\bf TS500} \\
   \cmidrule[0.3pt] (lr){2-4} \cmidrule[0.3pt] (lr){5-7}
     & Perplexity $\downarrow$   & Recovery \% $\uparrow$  & Worst \% $\uparrow$  & Perplexity $\downarrow$   & Recovery \% $\uparrow$  & Worst \% $\uparrow$ \\
   \midrule

   

    DenseCPD & - & 50.71 & - & - & 55.53 & - \\
    GraphTrans & 5.40 & 43.89 & 26.92 & 4.98 & 45.69 & 0.05 \\
    GVP &  4.71  &   44.14 &  33.73 & 4.20 &49.14 &  0.09  \\
    ProteinMPNN   &  3.93  &   54.43  &  37.24 &  3.53  & 58.08  &  0.03   \\
    PiFold  \   &  3.86  &   58.72  &  37.93 &   3.44 &60.42 &  0.03 \\ 
    \midrule

   ProteinMPNN+CMLM   & 3.60 (3.62) & 54.84 (54.22)  & 36.55 (41.18)    &3.46 (3.27) &57.44 (57.23) &\textbf{0.04} (0.04)    \\ 
   PiFold  (reimpl.) & 4.10 (3.70) & 53.74 (59.68) & 37.28 (38.14) & ~~n/a (3.70) & 58.82 (59.95) & \textbf{0.04} (\textbf{0.05})    \\
   \method (Pro.MPNN)   & 3.82 (3.60) & 56.92 (58.13) & 35.17 (39.14)  & \textbf{2.13} (\textbf{2.15})  & 64.30 (63.76) & 0.04 (0.04)  \\ 
   \method (PiFold)  & \textbf{3.50} (\textbf{3.27}) & \textbf{57.89} (\textbf{61.38}) & \textbf{39.74} (\textbf{46.75})   &3.19 (3.09) & \textbf{67.78} (\textbf{66.56}) & 0.02 (0.04) \\

   \bottomrule
   \end{tabular}
   }
   \vspace{-2.5mm}
\end{table}

\begin{wrapfigure}[14]{r}{0.5\linewidth}
    \vspace{-4mm}
    \centering
    \!\!\!\includegraphics[width=\linewidth]{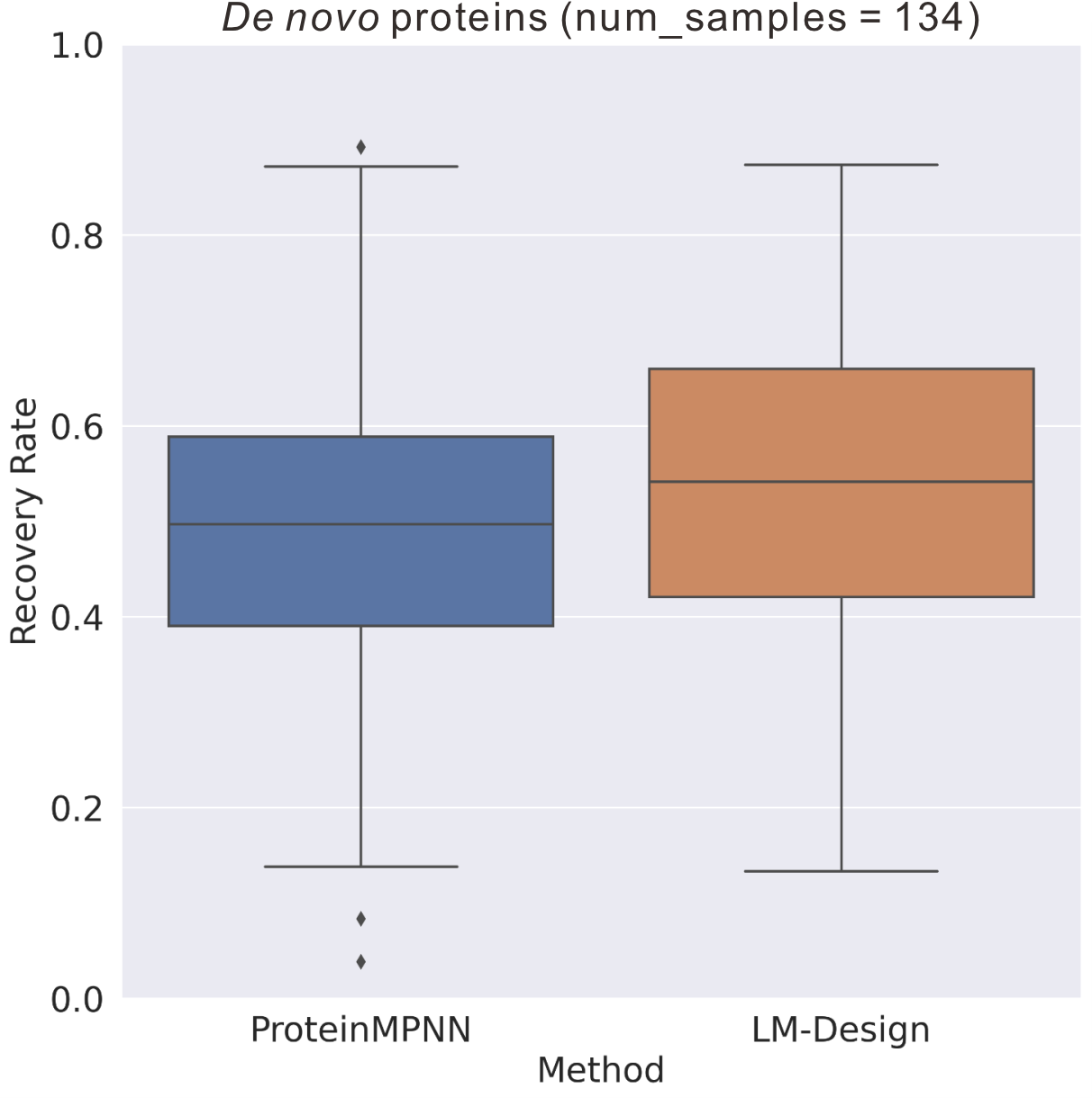}
    \vspace{-12pt}
    \caption{\method shows a better generalization capability on \textit{de novo} proteins than ProteinMPNN.}
    \label{fig:denovo}
\end{wrapfigure}
\paragraph{\textit{De novo} proteins.}
\textit{De novo} protein design explores the full protein sequence space for a new range of folds with high stability. In the last decade, advanced computational methods have been developed to design protein structures with atomic-level accuracy~\citep{huang2016coming}. To test whether \method can generalize to \textit{de novo} proteins, we compiled a \textit{de novo} protein dataset by collecting 134 \textit{de novo} protein monomers with different folds from Protein Data Bank. The performance of \method and ProteinMPNN was evaluated using this dataset showing an average recovery rate of 48.7\%. \method can recover the sequence with a significantly higher rate of 58.7\%, suggesting a better generalization capability on designed proteins. 
Please refer to \textbf{Fig.~\ref{fig:denovo}} for detailed results.

\begin{figure}[t]
    \centering
    \includegraphics[width=1\linewidth]{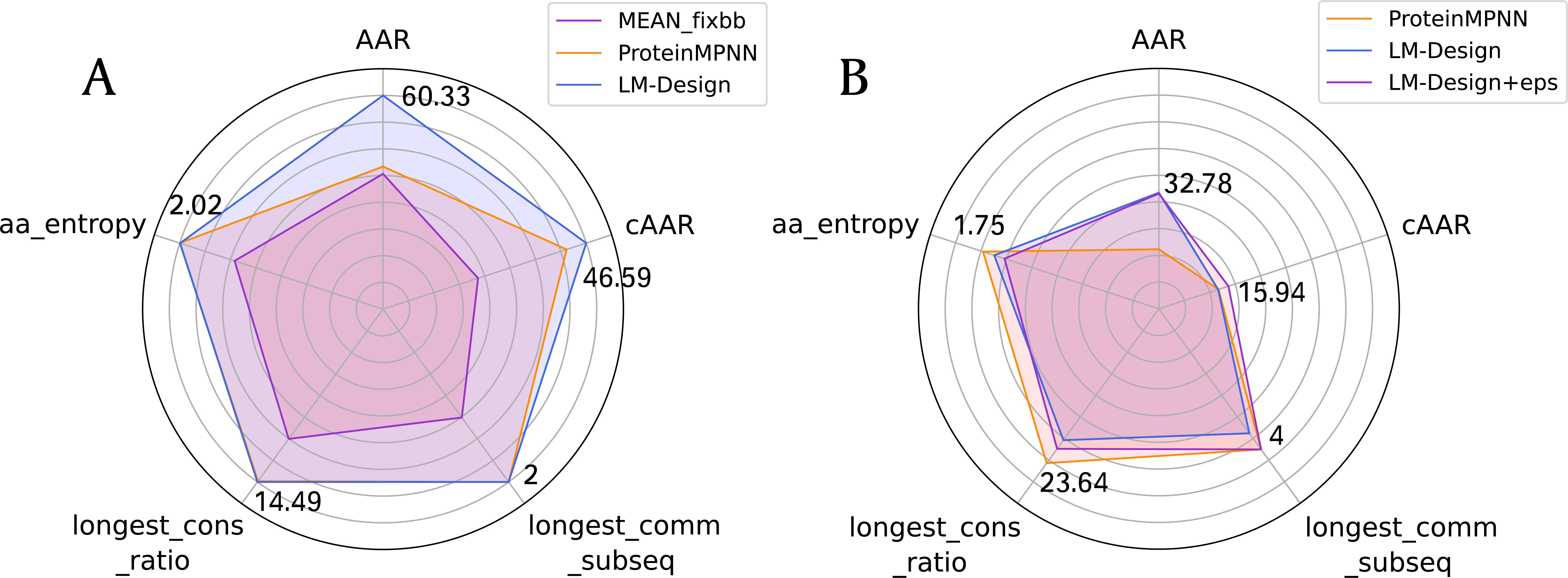}
    \vspace{-20pt}
    \caption{{\sl Performance comparison for antibody design.} The farther away from the center of the circle, the better the performance. (A) Antibody sequence design based on true structure; (B) Antibody sequence design based on predicted structure}
    \label{fig:antibody}
    \vspace{-5mm}
\end{figure}

\paragraph{Antibody design.}
Designing targeted antibodies for different antigens is one of the potential treatments for many diseases that currently cannot be cured. 
Antibody design is formulated as sequence infilling for complementary-determining regions (CDRs) given contexts (\ie, antigen and antibody frameworks). However, the commonly-used metric of sequence recovery (\textit{aka.} AAR, amino acid recovery) can be flawed as we often observed a ``mode collapse" problem for previous approaches due to extremely limited antibody data.
We hence design a package of five metrics regarding salient regions recovery, hallucinated sequence pattern, and entropy of predicted distribution for a more comprehensive evaluation (See Appendix~\S\ref{sec:antibody_metric} for details).
We conducted two kinds of experiments on RAbD dataset (\citet{adolf2018rosettaantibodydesign}) in designing CDR-H3 sequences given either true complex structures, or predicted ones.
(1) As for using true structures, we compare \method with ProteinMPNN and MEAN~\citep{kong2022conditional}, the SoTA neural antibody design approach using its \texttt{fixbb} mode.
As shown in \textbf{Fig.~\ref{fig:antibody}A}, \method outperforms antibody-specific MEAN model in terms of all evaluation aspects, showing that that models for general protein can effectively avoid mode failure while \pLMs help facilitate antibody design.
(2) In the case of predicted structures (\textbf{Fig.~\ref{fig:antibody}B}), the structures are predicted by MEAN.
The performance of \method and ProteinMPNN decreases significantly, implying they are fragile for predicted structures.
In contrast, \method + \texttt{eps}, where spatial perturbation is injected to the training structures~\citep{dauparas2022proteinmpnn}, shows stronger robustness hence better performance, suggesting the need of counter-adversarial considerations for structure-based sequence models to enhance generalizability.

\section{Discussions}
\label{sec:conclusion}

In this paper, we demonstrate that language models are strong protein designers, and introduce \method, a generic approach to enable \pLMs to yield preferable protein sequences for given folds.
We conduct a \textit{structural surgery} on \pLMs, where a lightweight structural adapter is implanted into \pLMs and endows it with structural awareness.
During inference, iterative refinement is performed to effectively optimize the generated protein sequences.
Experiments show that \method outperforms the state-of-the-art methods by a large margin.
Extensive analyses verify that \method can (1) effectively exploit both structural and sequential knowledge, (2) benefit from scaling data and size, and (3) generalize to other proteins (\eg, antibodies and \textit{de novo} proteins).

Despite a leap of accuracy in terms of native sequence recovery in this study, there remain several limitations. 
The ultimate goal of computational protein design is to completely design \textit{new} proteins based on a simple description of their target \textit{function}.
In machine learning, on the other hand, the dilemma of generalization (novelty) and memorization (recovery) is a long-standing question in evaluating generative models in general~\citep{van2021memorization}, namely, to what extent training data are memorized by the learning algorithm.
As a result, the most crucial and immediate concern is whether a high native sequence recovery is a good indicator for successful (structure-based) protein design, which has also been noted and discussed in recent literature~\citep{melnyk2022alphafold}.

For structure-based protein design, which lies in the current scope of this study, functions have already been determined by the given structures of interest\footnote{In the setting of fixed backbone sequence design, we assume structure design has been done beforehand as a prerequisite.}, the goal is therefore to explore protein sequence space to find novel sequences with optimal biochemical properties like stability and solubility.
In this case, we have already showcased that, albeit \pLMs being solely exposed to tens of millions \textit{native} protein sequences, \method is capable of leveraging the patterns learned by \pLMs and combining them to yield accurate yet diverse (\textbf{Fig.~\ref{fig:acc_diversity}}) and structurally valid (\textbf{Fig.~\ref{fig:structural_validation}}) protein sequences for given folds.

As for protein design at large, novel structures distant from natural proteins are often more emphasized~\citep{watson2022RFfold}. 
Recently, \citet{lin2022esmfold} show that sequential evolutionary knowledge learned by \pLMs corresponds deeply to protein structures and thus materializes protein structure prediction from single sequences.
In a concurrent work, \citet{verkuil2022language} further demonstrate that \pLMs at scale can synthesize \textit{de novo} proteins on the basis of the deep grammars learned from large-scale native sequences, generalizing beyond natural proteins, at a high experimental success rate.
Likewise, in the recent advances in generative AI algorithms in general, \eg, VAE~\citep{kingma2013vae}, autoregressive models~\citep{bahdanau2014neural,vaswani2017attention}, and diffusion probabilistic models~\citep{sohl2015diffusion,ho2020ddpm,rombach2022stablediffusion}, maximum likelihood estimation or its surrogates is the primary protocol for learning from massive native data by reconstruction. 
Once they have learned to sufficiently recover native data, they can synthesize completely new data, yielding impressive results and thus driving the recent surge of generative AI, both in academia and industry, and with great potential even in the field of protein (structure) design~\citep{watson2022RFfold}.

All of these show that, in our opinion, a good recovery is a prerequisite for novelty: a generative model with good adherence to the native data in its vastness and the ability to recover them (and their in-domain held-out set), if over-fitting does not occur, captures the underlying data distribution in a great extent, hence has the capability to generalize and synthesize \textit{de novo} data from the patterns and ``grammars'' they learned.

Based on the above discussions, we suggest that the good performance of \method can be attributed to the aforementioned reasons.
We expect \method can further benefit from more ingenious decoding/sampling algorithms for better \textbi{designability} (\eg, being more diverse, constrained, controllable, and programmable), integrating with structure prediction models~\citep{jumper2021AF2,baek2021RF} for the purpose of end-to-end protein \textbi{structure-sequence co-design}~\citep{jin2021iterative}, incorporating stochastic dynamics and variational Bayes to \textbi{better capture multimodal distributions} as in VAEs/diffusion models, and even utilizing protein analogs of \textbi{in-context learning}~\citep{wei2022emergent} or \textbi{chain-of-thoughts (CoT) prompting}~\citep{wei2022CoT,fu2022complexity} in LLMs for natural languages.
Moreover, as the scaling laws link performance with the computing, data, and model size in LLMs in general and proteins, we can be optimistic about the further breakthroughs in generative protein research enabled by the advances of scaled general sequence learning algorithms and their empowered protein applications, together with the progress in the fundamental protein geometry and structure learning.
We leave these exciting directions as future work.

\section*{Acknowledgements}
We would like to especially thank Dr. Hang Li for insightful discussions on the project and feedback on the manuscript that help shape this study. 
We thank Siyu Long, Xiaoqin Tan, and Yu Bao for their valuable comments.
We also thank Chloe Hsu for addressing our questions about the experimental evaluation in \citet{hsu2022esmif}.


\bibliography{references}
\bibliographystyle{icml2023}

\newpage
\appendix
\onecolumn
\section{Model Architecture of \method}
The overall architecture of \method is constituted by three components, \ie, a structure encoder, a \pLM as sequence decoder, and a structural adapter that bridges both. 
We directly utilized the established structure models (architectures, hyperparameters, \textit{etc.}) from literature to parameterize the structure encoder, in which we showcased ProteinMPNN~\citep{dauparas2022proteinmpnn} as our primary running example, and also investigated PiFold~\citep{gao2022pifold} and GVP-TransformerEncoder~\citep{hsu2022esmif} (we discarded the TranformerDecoder in GVP-Transformer and only kept the encoder part as our only purpose is encoding protein structure).
For \pLMs as the sequence decoder side, we primarily used the ESM-1b 650M~\citep{rives2019esm}, as well as ESM-2 series~\citep{lin2022esmfold}, with their pretrained model weights. 

Note that inventing better protein structure encoding modules or sequence pretraining/representation learning approaches is not the focus thus beyond the scope of our study. 
We built \method on top of these popular and publicly available models as backbones. 
We strongly encourage readers to check the details of these models in their original papers. 
In addition, we believe \method as a generic framework can further be improved by any future advances in the backbones in terms of both protein structure and sequence learning.

\begin{figure}[t]
    \centering
    \includegraphics[width=0.8\linewidth]{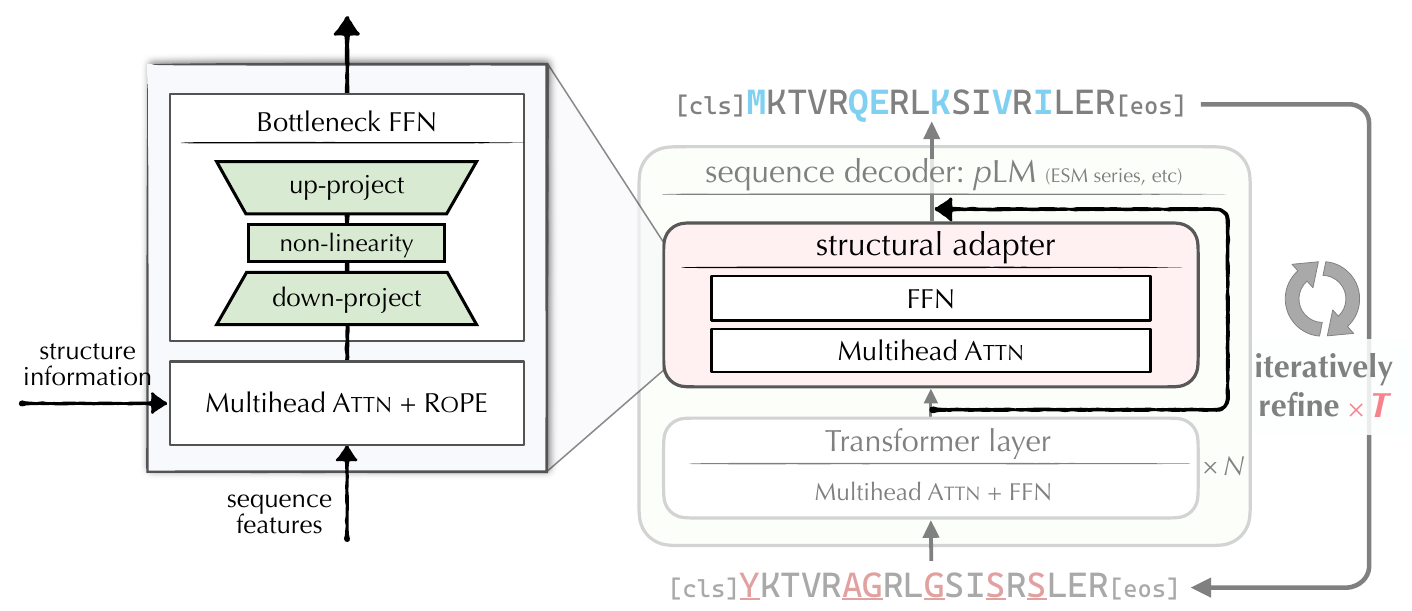}
    \caption{Illustration of our instantiation of the structural adapter}
    \label{fig:adapter}
    \vspace{-2.5mm}
\end{figure}

Here we elaborate on the design of the structural adapter. 
\textbf{Fig.~\ref{fig:adapter}} shows one instantiation we used in this paper, despite there possibly existing many others to explore, the structural adapter composes a multihead attention (\textsc{Multihead Attn}) that queries structure information from the structure encoder, followed by a \textit{bottleneck} feedforward network (FFN) to impose non-linearity and abstract features/representations. 
The bottleneck architecture of the FFN adheres to the best practice from \citet{houlsby2019parameter} and many follow-ups of it.
Most hidden dimensions of \textsc{Multihead Attn} and FFN are determined by the instance of structure encoder and \pLM we used, while the intermediate dimension of the bottleneck FFN was set to be half of the model dimension. 
\textsc{RoPe}~\citep{su2021rope} was used the supplement \textsc{Multihead Attn} for better modeling of positional information.
In all our experiments, only one structural adapter was placed after the last layer of \pLM, although layer-wise placement is also an obvious alternative and we leave it for future study.





\section{Additional Details on Benchmarking Evaluations}

\subsection{Single-chain Protein Design on CATH Datasets}
\paragraph{Datasets.}
We mainly compared \method against recent strong baselines on the CATH 4.2~\cite{orengo1997cath} dataset, using the same data splits as the compared systems\footnote{\url{https://github.com/jingraham/neurips19-graph-protein-design/blob/master/data/README.md}}, \eg, Structured Transformer~\citep{ingraham2019generative}, GVP~\cite{jing2020gvp}, ProteinMPNN~\citep{dauparas2022proteinmpnn}, and PiFold~\citep{gao2022pifold}, where proteins were partitioned by the CATH 4.2 topology classiﬁcation, resulting in 18024 proteins for training, 608 proteins for validation, and 1120 proteins for testing. 
To compare with ESM-IF~\citep{hsu2022esmif}, we also conducted evaluations on CATH 4.3\footnote{\url{https://github.com/facebookresearch/esm/blob/main/examples/inverse_folding/README.md}}, wherein 16153 structures are assigned to the training set, 1457 to the validation set, and 1797 to the test set.

\paragraph{Implementation, training, and metrics.}
The models were trained up to 100 epochs by default using the Adam optimizer on NVIDIA V100s. 
We used the same training settings as ProteinMPNN~\citep{dauparas2022proteinmpnn}, where the batch size was set to approximately 6000 residues, and Adam optimizer~\citep{kingma2014adam} with \texttt{noam} learning rate scheduler~\cite{vaswani2017attention} was used. 
We follow the previous studies to report perplexity and \textit{median} recovery scores on short proteins (length $\leq 100$), single-chain proteins (labeled with 1 chain in CATH, different from the single-chain concept which is in contrast with protein complex), and all proteins settings.

\subsection{Multi-chain Protein Complex Design}

\paragraph{Dataset and setting.}
We used the multi-chain complex dataset curated by \citet{dauparas2022proteinmpnn}, which collected protein assemblies in the PDB (as of Aug 02, 2021) determined by X-ray crystallography or cryoEM to better than $3.5\angstrom$ resolution and with less than 10,000 residues\footnote{\url{https://github.com/dauparas/ProteinMPNN/blob/main/training/README.md}}.
Sequences were clustered at 30\% sequence identity cutoff using \texttt{mmseqs2}~\citep{steinegger2017mmseqs2} resulting in 25,361 clusters. 
Those clusters were split randomly into three groups for training (23,358), validation (1,464), and testing (1,539), ensuring that none of the chains from the target chain or chains from the biounits of the target chain would be in the other two groups.
We kept the training setting the same as that of the single-chain scenario.




\section{Antibody Design}

\subsection{Task Definition, Datasets, and Evaluation Metrics}
\label{sec:antibody_metric}

\paragraph{Task definition.}
We design two types of experiments.
\begin{compactitem}
\item \textbf{Sequence design on true structures.}
In this experiment, we set the antibody sequence design task as a fixed-backbone protein design task in accordance with the training task of \method, wherein each model is asked to design the CDR-H3 region sequence of the antibody-based on the true antigen-antibody complex structure\footnote{CDRs, short for complementary-determining regions, are a few special regions on the antibody, with high variability and true interaction with the antigen, which determines the specificity of the antibody. CDR-H3 is the most important CDR.}.

\item \textbf{Sequence design on predicted structures.}
In this experiment, we examine the performance of the model when facing a more real situation, lacking the real structure of the antibody.
All the models are asked to predict the CDR-H3 sequence based on the predicted CDR structure. 
We use MEAN~\cite{kong2022conditional}, the SoTA method in the designing of antibody generate the structure of the CDR region.
Specifically, we modify the MEAN by initializing the CDR sequence using AbLang (an antibody pre-training model, \citet{olsen2022ablang}) to obtain a fine predicted CDR-H3 structure (RMSD $1.81 \rightarrow 1.70$).
\end{compactitem}

It should be noted that although antibodies are also proteins, antibody design, i.e. CDR design, is very different from protein design. Protein sequence diversity is evolutionary in origin, whereas CDR sequence diversity is derived from V(D)J recombination and somatic hypermutation in B cells, which indicates that CDR sequences are random to some extent.

\paragraph{Datasets.}
We follow the dataset adopted by most antibody design work. 60 antigen-antibody complexes from RAbD dataset were used to test \method's design capability on the antibody sequence when given the structure of the antigen-antibody complex, and all antibodies are renumbered under the IMGT scheme (\citet{lefranc2003imgt}).

\paragraph{Evaluation metrics.}
AAR is the most important metric in antibody design. However, because of the conservation of amino acids at both ends of the CDR region and the insensitivity of sequence recovery to short sequences, AAR may not reflect the true performance of the model.
For example, similar antibodies like ``\texttt{ARD G* Y* FDY}" are generated for different antigens, it fails to targeted design but still achieves good accuracy (37.5\% AAR on RAbD dataset). 

As a result, to evaluate the model's performance more comprehensively, we introduced four additional metrics to evaluate the generated CDR-H3 sequence in addition to widely used AAR. 
Concretely, \texttt{longest\_comm\_subseq} is the length of the longest common subsequence in at least 30\% of generated sequences, indicating the lack of diversity in the generated sequence, lower is better;
\texttt{longest\_cons\_ratio} is the proportion of the longest consecutive repeat sequence in the generated CDR sequence, indicating the occurrence of polyY and polyG in the generated sequence, lower is better.
\texttt{aa\_entropy} is the average entropy of each amino acid type in each generated CDR sequence, indicating the diversity of amino acid types in the CDR sequence, higher is better.
\texttt{cAAR} is the \texttt{AAR} calculated at the actual interaction position, which can better reflect the model's understanding of complex protein structures and interactions, higher is better.

\subsection{Additional Results}
\paragraph{Sequence design on true structures.}
MEAN is the method specially designed for co-designing CDR structures and sequences, and we used open-source code of MEAN and data (data were filtered down from 3127 to 2901 to remove the proteins that lacked atomic information) to retrain \texttt{fixbb} mode MEAN in this experiment, the training config is consistent with the original MEAN. 
The experiment results are shown in \textbf{Tab.~\ref{tab:antibody_design}}. 
In the first experiment, \method and ProteinMPNN+CMLM achieve far better performance than the original ProteinMPNN, and also outperformed MEAN in \texttt{fixbb} mode. 
Among all the five metrics, the performance gap between \method and ProteinMPNN+CMLM in \texttt{cAAR} is greater than the others. We believe that the language model in \method may ``forget" certain structural information, resulting in a significant decline in \texttt{cAAR}, which is closely related to structure. To avoid this ``forgetting" of the structure, we add the output logits of the structure encoder to \method's output logits to enhance the impact of the structure on the predicted sequence. Finally, \method+\texttt{Enc\_Logits} achieves the best performance in four of the five metrics.

\paragraph{Sequence design on predicted structures.}
In this experiment, 
The original ProteinMPNN achieves the best performance in three metrics of diversity, but shows great disadvantages in AAR (\textbf{Tab.~\ref{tab:antibody_design}}). \method and \method+Enc\_logits perform higher AAR, while ProteinMPNN+CMLM gets better cAAR. However, all methods show a huge performance degradation when using the predicted structure instead of the true structure, and this phenomenon shows the sensitivity of antibody design tasks to structural changes. We conducted preliminary experiments and try to mitigate this degradation, as we usually do not know the true structure of the CDR region in actual antibody design tasks.
Specifically, we inject spatial perturbation into \method's training structure and hope that this will help the model to be more robust when structures have uncertainty. The resulting model (\method+eps) achieves significant improvements in cAAR, proving that tolerance for structural bias is helpful for the actual antibody sequence design task.


\begin{table*}[t]
\centering
\small
\caption{{\sl Performance on RAbD dataset. The scores of the original ProteinMPNN were obtained using its publicly available model checkpoint.}}
\vspace{1.5pt}
\label{tab:antibody_design}
\resizebox{\linewidth}{!}{%
\begin{tabular}{llcccccc} 
\toprule
                                                                                        Structures & Metric & TRUE   & \method & \method+Enc\_logits                                     & ProteinMPNN+CMLM & ProteinMPNN      & MEAN\_fixbb      \\
\midrule
\multirowcell{5}{\rotatebox[origin=c]{0}{true}}                             & \textbf{longest\_comm\_subseq}                  & 2      & \multicolumn{1}{c!{\color{black}}}{\textbf{2}}    & 3  & 3           & \multicolumn{1}{c!{\color{black}}}{\textbf{2}}           & 6           \\
                                                                                        & \textbf{longest\_cons\_ratio}(\%)                   & 14.21      & 14.50    & \multicolumn{1}{c!{\color{black}}}{\textbf{12.60}} & 12.85           & 14.87           & 35.78           \\
                                                                                        & \textbf{aa\_entropy}                  & 2.01      & 2.02    & \multicolumn{1}{c!{\color{black}}}{\textbf{2.04}} & 2.00           & 2.03           & 1.48           \\
                                                                                        & \textbf{AAR}(\%)                    & 100      & 60.33    & \textbf{63.82}                                           & 61.51           & 40.26           & 38.20           \\
                                                                                        & \textbf{cAAR}(\%)                   & 100      & 46.59    & \multicolumn{1}{c!{\color{black}}}{\textbf{54.31}} & 51.88           & 42.10           & 21.80           \\
\midrule
                                                                                        &                                 & TRUE   & LM-Design & LM-Design+Enc\_logits                                     & LM-Design+eps    & ProteinMPNN+CMLM & ProteinMPNN      \\
\midrule
\multirowcell{5}{\rotatebox[origin=c]{0}{predicted}}                        & \textbf{longest\_comm\_subseq}                  & 2      & 5    & 5                                                    & \textbf{4}  & 6           & \textbf{4}           \\
                                                                                        & \textbf{longest\_cons\_ratio}(\%)                   & 14.21      & 35.00    & 30.44                                                   & 30.73  & 36.87           & \textbf{23.64}           \\
                                                                                        & \textbf{aa\_entropy}                  & 2.01      & 1.64    & 1.71                                                    & 1.54  & 1.54           & \textbf{1.75}           \\
                                                                                        & \textbf{AAR}(\%)                   & 100      & 32.78    & \begin{tabular}[c]{@{}c@{}}\textbf{34.22}\\\end{tabular} & 32.53           & 30.22           & 16.81           \\
                                                                                        & \textbf{cAAR}(\%)                   & 100      & 13.59    & 13.18                                                    & \textbf{15.94}  & 14.30           & 13.68           \\
                                                                                        
\bottomrule
\end{tabular}
}
\vspace{-10pt}
\end{table*}

In general, in antibody sequence design tasks, \method and ProteinMPNN+CMLM can greatly improve the accuracy of the generated sequence at the expense of limited diversity. More specifically, \method discards some perception of structure awareness based on ProteinMPNN+CMLM and gains a stronger ability to model sequences.

\section{\textit{De novo} Protein Design}
\paragraph{Task definition.}
The aim of computational protein design is to invent novel protein molecules with desired structures and functions~\citep{huang2016coming}. In this experiment, we evaluate the performance of the model to generalize to fixed-backbone sequence design on \textit{de novo} proteins. 

\paragraph{Dataset and evaluation metrics.} 
To test \method's generalization capability, we compiled 134 single-chain \textit{de novo} proteins (length $\leq 30$) from the PDB data bank. The structures of these samples were determined by X-ray crystallography or cryo-EM to better than 3.5Å resolution  The sequences were clustered at 30\% sequence identity cutoff using mmseqs2 ~\citep{steinegger2017mmseqs2}. We follow the other design tasks using median recovery scores for the test.

\section{Data Augmented Training with Predicted Structures from AlphaFold 2}
\label{appendix:B.5}
\paragraph{Task definition.}
Due to limited experimentally determined protein structure data and the surge of protein structure prediction models (\eg, AlphaFold 2), a natural idea for better protein inverse folding is to use the protein structure data predicted by AlphaFold 2 for data augmentation, which is similar to the back-translation in NLP area. For the protein design task, $\mathcal{X}$ and $\mathcal{Y}$ are the sets of protein structure coordinates and protein sequences. The goal is to learn a mapping $f: \mathcal{X} \to \mathcal{Y}$ while AlphaFold 2 has learned a mapping $g: \mathcal{Y} \to \mathcal{X}$. We denote those protein sequences without structures as $\mathcal{U}_{\mathcal{Y}} \subset \mathcal{Y}$. For any sequence $y_u \in \mathcal{U}_{\mathcal{Y}}$, we can predict its structure $\hat{x_u}=g(y_u)$ and add $(\hat{x_u}, y_u)$ to our training set.

\paragraph{Datasets.}
UniRef50~\citep{suzek2015uniref} is a sequence database that has over 50 million clusters at 50\% sequence identity, ESM-IF~\citep{hsu2022esmif} predicts structures of 12 million sequences in UniRef50 using AlphaFold 2. Since our main goal is to prove the feasibility of combining language model design and data augmentation, we use SWISS-PROT~\cite{boeckmann2003swiss} in our experiment. SWISS-PROT is a curated protein sequence database that strives to provide a high level of annotation, a minimal level of redundancy, and a high level of integration with other databases.

In order to prevent data leakage introduced by data augmentation, we need to exclude proteins that have the same fold as the proteins in validation and test splits. We annotated SWISS-PROT sequences with CATH fold classification according to the Gene3D~\cite{Lees2012Gene3D} database with \texttt{hmmsearch}~\citep{potter2018hmmsearch}. We then perform filtering based on all CATH folds from validation and test splits and randomly select a sub-set of 100, 000 examples from the filtered dataset. We fix the random seed and randomly select 20, 000 and 50, 000 sequences for different scales of data augmentation.


\section{Structural Contexts Analysis}
\subsection{Methods} 

In order to understand how \method improves protein design, we analyze the performance based on dissected structural contexts (folding core residues, surface exposed residues, residues located on loops, and residues located on complex interfaces). The dictionary of structural contexts is obtained by using the widely used DSSP ~\citep{kabsch1983dictionary} and \texttt{biopython}~\citep{cock2009biopython} tools. Solvent accessible surface area analysis from DSSP calculates the ratio of solvent accessible surface area to the maximum possible solvent accessible surface area for each residue. A cutting threshold of 0.1 is chosen to classify residues located in the folding core and on the exposed surface. DSSP is also used to provide secondary structure labels. Those residues showing turn, bend, and none secondary structure patterns are labeled as loops in this study. For interaction interface analysis, we utilize InterfaceBuilder from Biopython. When the structural context label is obtained for each residue, the average recovery rate is calculated on PDB structures.

\subsection{Performance Dissection Results}

\paragraph{Protein design.} 
As the results shown in \textbf{Fig.~\ref{fig:substructure}}. As expected, residues in the folding core with more structural constraints display better sequence recovery rates than surface exposed and loop residues using both the structure-based ProteinMPNN and sequence-based \method. Notably, the residues in the binding interface of multichain proteins show a poor recovery rate, suggesting the limited representation capability of existing structure-based methods in interaction interface design. \method can generally improve sequence recovery rates in different structural contexts, including the interaction interfaces.

\begin{figure}[t]
    \centering
    \includegraphics[width=0.7\linewidth]{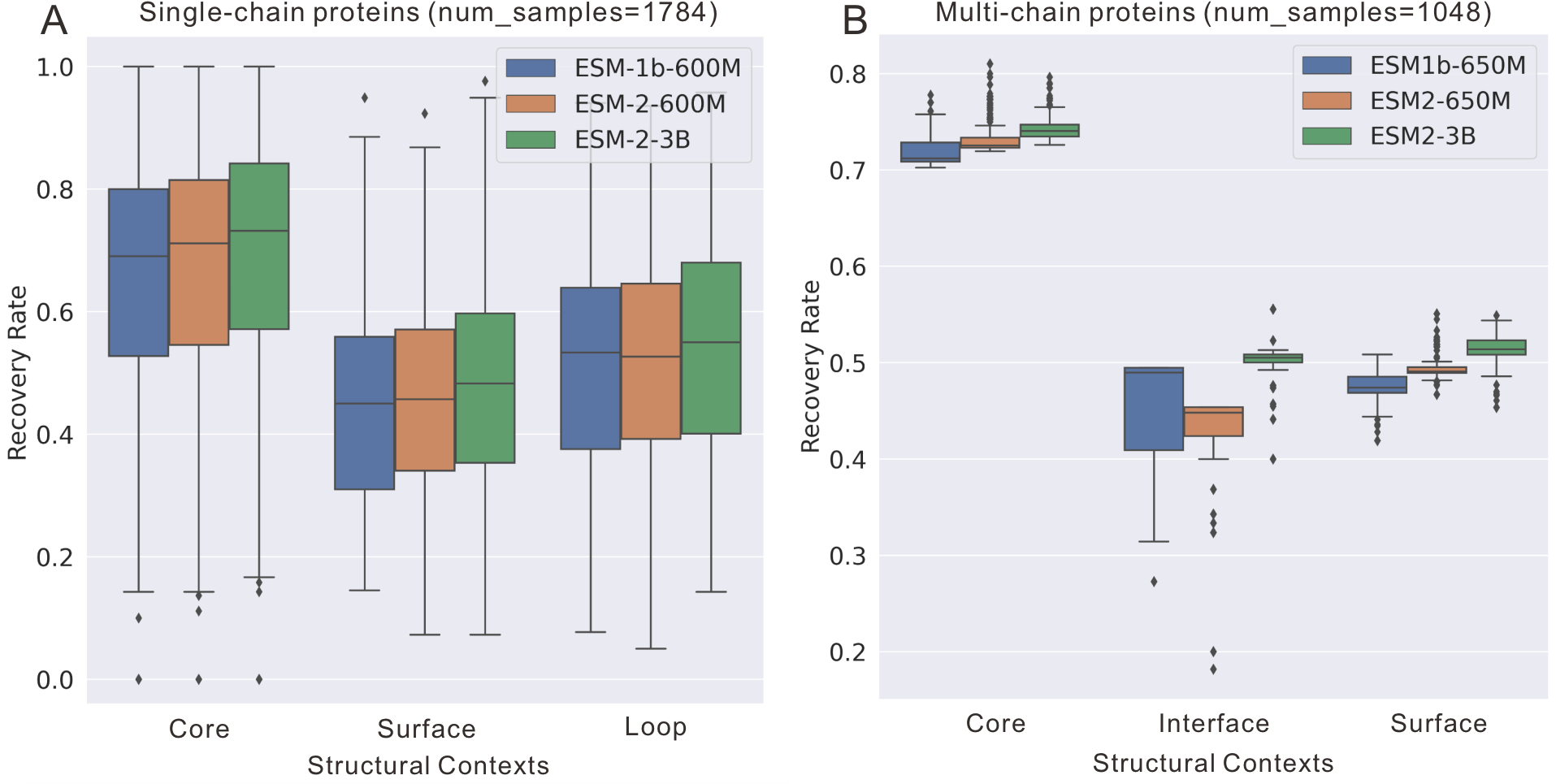}
    \caption{Comparison of sequence recovery by structural contexts for evaluation of the model scale of \textit{p}LMs. (A) Single-chain results. (B) Multiple-chain protein results.}
    \label{fig:LM_scale_substructure}
    \vspace{-2.5mm}
\end{figure}

\paragraph{\textit{p}LMs model scales.}
To study the impact of the model scale of \textit{p}LMs on single-chain and multi-chain protein design, we dissect their performance in different structural contexts. As shown in \textbf{Fig.~\ref{fig:LM_scale_substructure}}A, the recovery rate of residues in the folding core increased from 65.2\% (ESM-1b 650M) to 67.0\% (ESM-2 650M), and finally to 69.3\% using ESM-2 3B. For multi-chain proteins, ESM-2 3B also achieves the best recovery rate among different structure contexts (\textbf{Fig.~\ref{fig:LM_scale_substructure}}B). Therefore, the performance improvement can be found as a general trend for different structural contexts in sequence recovery when using \pLMs with a larger scale of parameters. This result suggests larger \pLMs models benefit from the training on the larger-scale protein sequence.

\section{Related Work}
\label{sec:related_work}

\paragraph{Structure-based Protein Sequence Design with Deep Generative Models.}
Deep generative modeling typically formulates structure-based protein sequence design as a conditional sequence generation problem, wherein protein 3D structures can typically be represented as a \textit{k}-NN graph~\citep{ingraham2019generative}. 
Several graph neural networks (GNNs) can be applied in this case to derive protein structural features. The protein graph establishes edge features between adjacent residues and encodes residue information as node features. The graph attention encoder and autoregressive decoder are used by GraphTrans~\citep{ingraham2019generative} for protein design. The novel geometric vector perceptrons which take into account both scalar and vector features, GVP~\citep{jing2020gvp} improve further performance. Global graph attention for protein design is further introduced by GCA~\citep{tan2022generative} to capture long-range information.
Recently, ProteinMPNN~\citep{dauparas2022proteinmpnn} and PiFold~\citep{gao2022pifold} introduce more complicated protein features and expressive GNNs and gain significant improvements. 
Besides the primary generative purpose, \citet{yang2022maskedif} attempt to use this task as a proxy for protein (structure-aware) representation learning.
In this study, we develop \method on top of the powerful structural capability of ProteinMPNN and PiFold, while \method can further benefit from future progress in deep geometric learning for proteins. 

There, however, remains a crucial and severe issue that limits deep generative models for structure-based protein design is the lack of sufficient protein structure data.
To address this, ESM-IF~\citep{hsu2022esmif} achieves dramatic improvements with 
effective data augmentation. 
\citet{hsu2022esmif} augments 16k experimental structures from CATH with 12 millions additional predicted structures ``back-translated''~\cite{sennrich2015improving} by AlphaFold 2~\cite{jumper2021AF2}. 
By further scaling the model to over a hundred million parameters, the ESM-IF model~\cite{hsu2022esmif} exhibits the state-of-the-art performance by a gain of over 10\% sequence recovery (\ie, 51.6\% on CATH 4.3), compared to the scenarios where only experimental data is considered. 
Note that \method is orthogonal to such an effective data augmentation, which has been empirically verified by our experiment.
By combining our modeling advances with data augmentation, we can take the best of both worlds to push the boundary of structure-based protein design.

\paragraph{Protein Language Models.}
There is growing interest in developing protein language models (\pLMs) at the scale of evolution due to the abundance of 1D amino acid sequences, such as the series of ESM~\citep{rives2019esm,lin2022esmfold}, TAPE~\citep{rao2019evaluating}, 
ProtTrans~\citep{elnaggar2021prottrans}, PRoBERTa~\citep{nambiar2020transforming}, PMLM~\citep{he2021pre}, 
ProteinLM~\citep{xiao2021modeling}, 
PLUS~\citep{min2021pre}, 
Adversarial MLM~\citep{mcdermott2021adversarial}, 
ProteinBERT~\citep{brandes2022proteinbert}, 
CARP~\citep{yang2022convolutions} in masked language modeling (MLM) fashion, 
ProtGPT2~\citep{ferruz2022protgpt2} in causal language modeling fashion, and several others~\citep{melnyk2022reprogramming,madani2021deep,unsal2022learning, nourani2021tripletprot, lu2020self, sturmfels2020profile, strodthoff2020udsmprot}.
These protein language models are able to generalize across a wide range of downstream applications and can capture evolutionary information about secondary and tertiary structures from sequences alone.
They have recently been demonstrated with strong capabilities in uncovering protein structures~\citep{lin2022esmfold}, predicting the effect of sequence variation on function~\citep{meier2021language}, antibody infilling~\citep{melnyk2022reprogramming} and many other general purposes~\citep{rives2019esm}.

Concurrently, \citet{verkuil2022language} demonstrate that \pLMs at scale can generalize beyond natural proteins to generate \textit{de novo} proteins, and validate their hypothesis \textit{in silico} and experimentally in great detail, in which \pLMs are capable of even designing protein structure even though they only trained on sequences. 


\paragraph{Parameter-efficient Tuning for Language Models in NLP.}
The ability to effectively tune and adapt LLMs with few trainable parameters is crucial for modern NLP research and applications, because it is becoming increasingly impractical to train and store full copies of large LLMs for a variety of downstream tasks.
Based on whether new trainable parameters are added, existing methods for LLM adaptation can be roughly divided into two categories.
For example, one could train a fraction of the model's parameters, such as the prediction head and bias terms~\citep{lee2019would} or introduce task-specific parameters to various LLM components~\citep{li2021prefix}.
Among them, \citet{houlsby2019parameter} have studied the choice of adapter architectures and the placement of them within LLMs in great detail. They discover that a stack of bottleneck networks, which only adds a few extra parameters to the network, performs well. The majority of the subsequent research~\citep{pfeiffer2020adapterfusion,pfeiffer2020mad} and community-contributed framework, the AdatperHub~\citep{pfeiffer2020adapterhub}, are inspired by this design.

However, there yet remains less effort in transferring protein large language models for downstream important protein problems of interest, especially for the generative setting such as protein design. 
To the best of our knowledge, \method is among the first attempts to steer \pLMs to perform protein design, surpassing all previous non-\pLMs empowered approaches for this task, suggesting the call for more attention in the power and potential of \pLMs for protein research. 


\end{document}